\documentclass[runningheads]{llncs}
\usepackage[T1]{fontenc}
\usepackage{graphicx}
\usepackage{booktabs}
\usepackage[misc]{ifsym}
\newcommand{\corr}{(\Letter)}
\usepackage{mwe}

\usepackage{multirow}
\usepackage{amsmath,amssymb,amsfonts}
\usepackage{mathrsfs}
\usepackage[title]{appendix}
\usepackage{xcolor}
\usepackage{textcomp}
\usepackage{manyfoot}
\usepackage{algorithm}
\usepackage{algorithmicx}
\usepackage{algpseudocode}
\usepackage{listings}
\usepackage{verbatim}
\usepackage{graphicx}
\usepackage{caption}
\usepackage{subcaption}
\usepackage{booktabs}
\usepackage{hyperref}
\usepackage{cleveref}
\usepackage{array}
\usepackage[para]{footmisc}
\usepackage{longtable}
\usepackage{tikz}

\newcommand{\feasibilityGraph}{G_U}
\newcommand{\vectorx}{\mathbf{x}}

\newcommand{\vectorxi}{\mathbf{x_i}}
\newcommand{\vectorxj}{\mathbf{x_j}}

\begin{document}

\title{FACEGroup: Feasible and Actionable Counterfactual Explanations for Group Fairness
\thanks{This paper has been accepted for publication at the ECML PKDD 2025 conference.}
}
\titlerunning{Feasible and Actionable Counterfactual Explanations for Group Fairness}

\author{Christos Fragkathoulas\inst{1,2}\corr \and
Vasiliki Papanikou\inst{1,2} \and
Evaggelia Pitoura\inst{1,2} \and
Evimaria Terzi\inst{2,3}
}
\institute{University of Ioannina
\email{pitoura@uoi.gr}
\and
Archimedes, Athena Research Center, Greece \email{\{ch.fragkathoulas, v.papanikou\}@athenarc.gr}
\and
Boston University, USA \email{evimaria@bu.edu}}
\authorrunning{C. Fragkathoulas et al.}

\tocauthor{Christos Fragkathoulas, Vasiliki Papanikou, Evaggelia Pitoura, Evimaria Terzi}
\toctitle{FACEGroup: Feasible and Actionable Counterfactual Explanations for Group Fairness}

\maketitle

\begin{abstract}
Counterfactual explanations assess unfairness by revealing how inputs must change to achieve a desired outcome.
This paper introduces the first graph-based framework for generating group counterfactual explanations
to audit group fairness, a key aspect of trustworthy machine learning. 
Our framework, FACEGroup (Feasible and Actionable Counterfactual Explanations for Group Fairness), models real-world feasibility constraints, identifies subgroups with similar counterfactuals, and captures key trade-offs in counterfactual generation, distinguishing it from existing methods. 
To evaluate fairness, we introduce novel metrics for both group and subgroup level analysis that explicitly account for these trade-offs.
Experiments on benchmark datasets show that FACEGroup effectively generates feasible group counterfactuals while accounting for trade-offs, and that our metrics capture and quantify fairness disparities.

\end{abstract}

\section{Introduction}
AI-driven technologies increasingly shape critical decisions, making it essential to understand their underlying reasoning and evaluate their fairness. A variety of explanation methods have been proposed to enhance transparency ~\cite{fragkathoulas2024explaining,dwivedi2023explainable}, with counterfactual explanations (CFs) gaining prominence~\cite{10.1145/3677119}.
 Individual CFs reveal how modifying specific features can alter model decisions, offering actionable insights. For example, consider a person whose loan application is rejected by a machine learning model; a CF might indicate that increasing annual income or reducing the debt-to-income ratio would lead to approval.

Prior work has primarily focused on individual counterfactual explanations (CFs) ~\cite{kanamori2020dace,guyomard2023generating,fragkathoulas2025ugce,ustun2019actionable,poyiadzi2020face,ribeiro2016should,mothilal2020explaining,10.1145/3375627.3375812,bynum2023counterfactuals,karimi2020model}, with comparatively few studies addressing counterfactuals for groups of instances ~\cite{rawal2020beyond,ley2023globe,kanamori2022counterfactual,kavouras2024fairness}. 
Group counterfactual explanations (GCFs) identify how a group of instances, often defined by shared characteristics or \textit{protected attributes} such as sex or race, could collectively alter their features to achieve favorable outcomes. GCFs are not simply aggregations of individual CFs; rather, they reveal common patterns or barriers affecting the group as a whole, which is critical for understanding systemic disparities and informing policy or organizational decisions.
Previous studies introduce group-based approaches, by identifying common patterns among individuals with favorable outcomes~\cite{rawal2020beyond}, learning global translation vectors, and scaling them for GCFs~\cite{ley2023globe}, or constructing decision trees via stochastic local search~\cite{kanamori2022counterfactual}. In contrast, our work is the first to generate GCFs using a graph-based approach that enforces feasibility, supports subgroup-level analysis, and explicitly addresses the key trade-offs involved in counterfactual generation.

FACEGroup, our approach for generating Feasible and Actionable Group Counterfactual Explanations (GCFs), generates GCFs using a density-weighted feasibility graph~\cite{poyiadzi2020face}, where nodes represent data points and 
edges denote feasible transitions that comply with real-world constraints. To ensure plausibility, we restrict connections to allow only small feature changes between data points.
A key property of this graph is that feasibility constraints, cost limitations, and density weighting naturally partition the data into weakly connected components (WCCs), effectively dividing each group into subgroups with similar feasible counterfactual explanations.

The generation of group counterfactual explanations (GCFs) inherently involves balancing several key trade-offs: the proportion of factual instances within a group that are explained by the selected set of counterfactuals (coverage), the effort or change required for group members to achieve a counterfactual (cost), and the number of unique counterfactuals generated for the group (interpretability). To address these trade-offs, we introduce two algorithmic formulations based on the feasibility graph: the cost-constrained approach, which maximizes group coverage under a cost limitation, and the coverage-constrained approach, which minimizes the maximum cost required to achieve a specified coverage level. Both formulations are supported by mixed-integer programming solutions and greedy heuristics that operate at both the group and subgroup levels. Our approach also ensures that the generated counterfactuals remain feasible and actionable.

Finally, we introduce novel fairness metrics for group counterfactuals, which enhance existing fairness measures by capturing the various trade-offs in counterfactual generation and can be applied at both group and subgroup levels.
We evaluate FACEGroup on real-world datasets, showing its effectiveness in fairness auditing.
Compared to existing methods, FACEGroup produces 
more feasible and compact counterfactuals that 
align with the data distribution.

The rest of this paper is structured as follows: Section~\ref{seq: problem_def} formalizes the problem, Section~\ref{seq: algorithms} presents our algorithms, Section~\ref{seq: fairness} introduces our fairness measures, Section~\ref{seq: experiments} details experiments, Section~\ref{seq: related} discusses related work, and Section~\ref{seq: conclusion} concludes.

\section{Problem Definition}
\label{seq: problem_def}
Let $ f: \mathbb{R}^d \rightarrow \{0,1\} $ be a binary classifier which maps instances in a $d$-dimensional feature space into two classes, labeled 0 and 1. Let $ U \subseteq \ \mathbb{R}^d$ denote the input space.
A model prediction on an individual instance $\mathbf{x}$ $\in$ $U$, called \textit{factual}, is explained by crafting a counterfactual (CF) instance $\mathbf{x'}$ $\in$ $\mathbb{R}^d$ that is similar to $\mathbf{x}$ but leads to a different outcome, i.e., $f(\mathbf{x'}) \neq  f(\mathbf{x})$ \cite{10.1145/3677119}. The changes in feature values from $\mathbf{x}$ to $\mathbf{x'}$ should be feasible and comply with real-world constraints, for instance, changes to immutable features, such as race or height, should be prohibited.
Formally, a counterfactual $\mathbf{x'}$ for $\mathbf{x}$ is defined as: $\vectorx' = {\mathrm{arg}}\ {\mathrm{min_{\vectorx'' \in \mathcal{A}_\vectorx}}\ cost(\vectorx, \vectorx'') \, \text{s.t.} \,
f(\vectorx'') \neq f(\vectorx)},$
where $cost(\vectorx, \vectorx'')$ is a function measuring the cost of transitioning from $\vectorx$ to $\vectorx'$. The \textit{feasibility set} $\mathcal{A}_\vectorx$ denotes the set of counterfactuals attainable from $\mathbf{x}$ via feasible changes.

It would be hard to trust a CF if it resulted in a combination of features that were unlike any observations the classifier has encountered before \cite{10.1145/3677119}. Therefore, CFs should also be coherent with the underlying data distribution.
To ensure both feasibility and plausibility, we adopt a graph-based approach.
Following \cite{poyiadzi2020face}, we construct a weighted directed graph $\feasibilityGraph = (V, E, W)$. 
Nodes correspond to instances in \( U \), and an edge from node \( \mathbf{x_i} \) to node \( \mathbf{x_j} \) represents a feasible transition in the feature space.  
We call this graph \textit{feasibility graph}.
Transitions are further constrained by a cost threshold $\epsilon$, ensuring that only small-cost feature changes are allowed.
This ensures that changes between instances are both feasible and small. 
The weight function $W$ is defined using a density-based approach \cite{poyiadzi2020face} to ensure that CFs lie in dense areas of the input space and avoid outliers. Each edge in \( \feasibilityGraph \) is assigned a weight \( W_{ij} \), calculated as the product of the density of the instances around the midpoint of \( \mathbf{x_i} \) and \( \mathbf{x_j} \) estimated using a Kernel Density Estimator (KDE) \cite{10.1214/aoms/1177728190}, and the cost between instances:
\(
    W_{ij} = KDE\left(\frac{\mathbf{x_i} + \mathbf{x_j}}{2}\right) cost(\mathbf{x_i}, \mathbf{x_j}).
\)

Given $\feasibilityGraph$, we now formally define the feasibility set $\mathcal{A}_\vectorx$ of factual $\vectorx$ as the set of instances $\vectorx'$ for which there is a path in $\feasibilityGraph$ from $\vectorx$ to $\vectorx'$, i.e., the set of instances that are reachable from $\vectorx$:
$\mathcal{A}_\vectorx = \{\vectorx' \in \mathbf{U} | \vectorx'$  is reachable from $\vectorx \text{ in } \feasibilityGraph\}$. These instances are the \textit{feasible} CFs for $\vectorx$.

Instead of finding a CF for a single factual $\mathbf{x}$, we are interested in providing CFs
for a set $X \subseteq U$ of instances mapped to the same class.
Let $X' \subseteq U$ be the set of instances mapped to the opposite class.
Our goal is to identify a small subset $S$ of $X'$ of size $k$ that best explains $X$. 
We limit the number of CFs to $k$ for interpretability.
To select $S$, we consider coverage-cost trade-offs.
For a set of CFs $S \subseteq X'$, coverage is:
 \[coverage(X, S) = \left| \{ \mathbf{x} \, | \, \mathbf{x} \in X \text{ and } \exists \text{ }  \mathbf{x'} \in S \cap \mathcal{A}_\vectorx\} \right|.\]
We overload the notation for $cost$ to define the cost between an instance and a set, as well as between two sets:
\[cost(\vectorx, S) = \min_{\vectorx' \in S} cost(\vectorx, \vectorx'), \quad cost(X, S) =
    \min \max_{\vectorx \in X} cost(\vectorx, S).\]
The function $cost(\mathbf{x}, \mathbf{x'})$ captures the cost of transforming $\mathbf{x}$ to $\mathbf{x'}$, offering flexibility to adapt to specific problem requirements. 
For example, cost can be defined as the vector distance (e.g., L2 norm), the sum of edge weights along the shortest path in $\feasibilityGraph$, or simply the number of hops on this path.
By emphasizing proximity in feature space and by considering dense paths, these definitions ensure that the CFs are closely aligned with the data distribution.
Our approach works with any definition of cost.

A necessary condition for $\mathbf{x}'$ to be a feasible counterfactual for $\mathbf{x}$ is that both $\mathbf{x}$ and $\mathbf{x}'$ belong to the same weakly connected component (WCC) of $\feasibilityGraph$. As a result, $\feasibilityGraph$ induces a partition of the set of factual instances $X$ into $m$ disjoint subsets $X_1, \dots, X_m$,  $m > 0$. Each subset $X_i$ contains instances in $X$ that belong to the same WCC of $\feasibilityGraph$ and thus share a common space of feasible counterfactuals, denoted $X'_i$, which also reside within the same component.
This partitioning of $X$ into subgroups with distinct feasible counterfactual spaces offers a meaningful perspective for analyzing model behavior at both the group and subgroup level, highlighting regions of the input space that support similar feasible explanations.

We now provide two definitions of the FACEGroup problem. Our first definition prioritizes cost over coverage, setting a threshold on cost, and our second definition prioritizes coverage over cost, asking for a set that provides a specified coverage degree $c$.
\begin{problem} [Cost-Constrained]
 Given $X$, $X'$, $k$ $\in \mathbb{N}^{*}$, and cost threshold $d \in \mathbb{R}_{*}^+ $, find $S\subseteq X'$ with $|S|\leq k$ and $Q\subseteq X$ such that for every instance $\vectorx\in Q$ there exist an instance $\vectorx'\in S$ such that $cost(\vectorx, \vectorx') \leq d$ and $|Q|$ is maximized.
\end{problem} 
\begin{problem} [Coverage-Constrained]
Given $X$, $X'$, $k$ $\in \mathbb{N}^{*}$,  and coverage degree $c$, $0 < c \leq 1$, find $S \subseteq X'$ with $|S|\leq k$ such that $coverage(X, S) \geq c \, |X|$ and $cost(X,S)$ is minimized.
\end{problem}

\section{Algorithms}
\label{seq: algorithms}
Our approach to generating feasible CFs is based on the feasibility graph $\feasibilityGraph$.
Both optimization problems are NP-hard. The cost-constrained problem can be formulated as an instance of the maximum coverage problem, while the coverage-constrained problem is similar to the classical $k$-center problem \cite{tansel1983location}.

In the following, we present two versions for both problems: (a) a global version that generates CFs for the whole set $X$ and (b) a local version that generates CFs per subgroup $X_i$. We also show how the local version can be used to generate CFs for the whole group $X$.
A common step in both problems involves computing, for each factual $\mathbf{x}$, the candidate counterfactuals, i.e., the feasibility set $\mathcal{A}_{\mathbf{x}}$ and computing costs.
To this end, we use Breadth-First-Search for vector costs (e.g., L2 distance) and Dijkstra's algorithm for shortest path costs, with complexities of $O(|V| + |E|)$ and $O(|V|\log|V| + |E|)$, respectively.

\subsection{The Cost-Constrained FACEGroup Problem}\label{sec:cost-constrained}
We solve this problem using two approaches: (a) a Mixed-Integer Programming (MIP) that explicitly models constraints for each factual-counterfactual pair while optimizing coverage, and (b) a Greedy approach that iteratively selects CFs to maximize coverage.

For the MIP solution of the global version of the problem, we define two binary decision variables. Let $r_{\vectorx\vectorx'} = 1$ if $\vectorx'$ covers $\vectorx$; and $r_{\vectorx\vectorx'} = 0$, otherwise, and $u_{\vectorx'} = 1$ if CF $\vectorx'$ covers any instance in $X$, and $u_{\vectorx'} = 0$ otherwise.
The goal is to maximize the number of covered factual instances:
\begin{gather}
  \max \sum_{\vectorx' \in X'} \sum_{\vectorx \in X} r_{\vectorx\vectorx'} 
\quad \text{s.t.}\quad \sum_{\vectorx' \in X'} u_{\vectorx'} \le k \quad (1) \notag\\
\sum_{\vectorx' \in X'} r_{\vectorx\vectorx'} \le 1,\ \forall \vectorx \in X \quad (2)
\quad r_{\vectorx\vectorx'} \le u_{\vectorx'},\ \forall \vectorx'\in X', \forall \vectorx\in X \quad (3), \notag\\
\quad  u_{\vectorx},\,r_{\vectorx\vectorx'} \in \{0,1\},\ \forall \vectorx\in X,\ \vectorx'\in X'.
\quad (4) \notag
\end{gather}
While constraint ($1$) limits the number of selected CFs to at most $k$, constraint ($2$) enforces that each factual instance $\vectorx$ is assigned to at most one CF  $\vectorx'$. Constraint ($3$) guarantees that if a CF $\vectorx'$ is assigned to cover a factual instance $\vectorx$ ($r_{\vectorx\vectorx'}=1$) then  $\vectorx'$ must be selected $u_{\vectorx'}=1$,
and constraint ($4$) defines the binary decision variables. This formulation has $O(2^{|X'|})$ complexity.

For the global Greedy version of the problem, we iteratively select counterfactuals (CFs) to maximize coverage. 
Let $S_t$ be the set of counterfactuals selected at iteration $t$.
We start with an empty set \( S_0 = \emptyset \).
At each iteration \( t \), the algorithm selects the CF \( \mathbf{x'} \in X' \) that 
\begin{align*}
\vectorx' = \arg\max_{\mathbf{\vectorx''} \in X'} (coverage(X, S_{t-1}) + coverage(X, \{\vectorx''\}), \tag{5}
\label{alg: greedy_max_cov}    
\end{align*}
updates \( S_t = S_{t-1} \cup \{\mathbf{x'}\} \), and terminates when either \( |S_t| = k \) or all instances in \( X \) are covered.

The worst-case complexity of this algorithm is $O(k |X|)$. Given the submodular nature of coverage, where the marginal gain of adding a new CF to the set $S$ decreases as $S$ grows, it adheres to the properties of submodular maximization. 
Consequently, the attained coverage is no worse than $(1 - \frac{1}{e})$ times the optimal maximum coverage \cite{hochba1997approximation}.

The Greedy algorithm can also be used to provide a counterfactual explanation for a subgroup $X_i$ by applying it only to the corresponding WCC. We can also utilize this local version to provide counterfactuals for the whole group $X$ by applying the Greedy algorithm iteratively to all $m$ WCC as follows. Initially, we apply a single step of the Greedy algorithm at each WCC. Then, we select the CF that provides the best coverage and apply an additional step of the algorithm to the WCC from which the CF was selected. We repeat this until the maximum number $k$ of counterfactuals is reached or all factual instances are covered. It is easy to see that this local version provides the same result as the global one.
The local Greedy selection has the same complexity as the global Greedy approach, as it follows a similar process while iterating over WCCs, either scanning all $|X'|$ candidates or evaluating coverage within each component.

\subsection{The Coverage-Constrained FACEGroup Problem}
To solve this problem, we employ two algorithms: a mixed-integer programming (MIP) and a Greedy 2-approximation algorithm \cite{gonzalez1985clustering}. While the Greedy algorithm provides an efficient yet approximate solution, the MIP guarantees optimal results \cite{daskin1997network}, but can become computationally expensive for large graphs.

For the MIP formulation, the solution is similar to the Cost-Constrained problem with the following modifications. The objective function minimizes the maximum cost $d$ of the farthest instance while ensuring that $coverage(X, S) \geq c \,|X|$. Constraints ($1$), ($2$), ($3$), and ($4$) still apply, along with:
\begin{align*}
\sum_{\vectorx' \in X'}cost(\vectorx,\vectorx')r_{\vectorx\vectorx'} \leq d, \;\; \forall \vectorx \in X
\quad \text{($6$),} & \quad
\sum_{\vectorx' \in X'} \sum_{\vectorx \in X} r_{\vectorx\vectorx'} \geq c \, |X| \quad \text{($7$).} 
\end{align*}

Constraint ($6$) ensures that the cost of any node to its assigned center does not exceed $d$, enforcing the objective function, and Constraint ($7$) enforces that the desired coverage percentage is achieved. For full coverage, $c = 1$, constraint ($2$) becomes an equality constraint, and constraint ($7$) is no longer needed.

For the Greedy algorithm, the process begins by arbitrarily selecting the first counterfactual $\vectorx'$ and assigning all factuals $\vectorx$ within a cost of $r$ to it, where $r$ is initially set to the maximum cost between any factual and candidate counterfactual. We then iteratively select the counterfactual that is farthest from those already chosen and assign all factuals within a cost of $r$ to it. This process continues until we reach the predefined coverage or the number of counterfactuals $k$. To find the smallest value of $r$ that satisfies the coverage requirement, we employ a binary search. The complexity of this algorithm is $O(k |X| log(d))$, since it assigns up to $|X|$ factuals for each of the $k$ selected counterfactuals and binary search adds this logarithmic factor $log(d)$, where $d$ is the range of costs considered. 

Both the MIP and the Greedy approaches can be applied globally and locally. In the global version, we apply the algorithms on the $G_U$ graph.
In the local version, for a specific subgroup $X_i$ of $X$, the algorithms are applied within the corresponding $WCC$ of $G_U$.
  
We now describe how the local version can be used to solve the global version.
\label{algo: Mips local to global full coverage} 
Consider the case of full coverage (\( c = 1 \)) with \( m \) WCCs ordered arbitrarily as \( C_1, C_2, \ldots, C_m \). Achieving full coverage reduces to distributing \( k \) counterfactuals among these components. Since at least one counterfactual is required per WCC, the maximum allocation per WCC is at most \( k - m \).  
First, we run MIP or Greedy within each WCC, varying \( k \) from 1 to \( k - m \). Let \( l_i \) be the minimum counterfactuals needed to fully cover \( C_i \). We start by assigning \( l_i \) to each \( C_i \), then iteratively allocate remaining counterfactuals to the WCC with the highest cost until the total reaches \( k \).

When $c < 1$, the task becomes more complex as we have to allocate both $k$ and coverage $c$ across the WCCs. Let $F({1...i},k, n)$ be the minimum cost of allocating $k$ counterfactuals that cover a total of $n$ factuals considering connected components $WCC_1,...., WCC_i$, where \( n \) $ = c |X| $.  
Similarly, let \( F(i, k, n) \) represent the minimum cost of allocating \( k \) counterfactuals to cover \( n \) factuals within component \( WCC_i \).
Then, we can solve the problem with time complexity of \( O(m (k n)^2) \), using dynamic programming as follows:
\begin{align*}
F({1...i}, k, n) = & \min_{1 \leq n' \leq n, 1 \leq k' \leq k} & \left\{ F({1...i-1}, k-k', n-n') + F(i, k', n') \right\}
\end{align*}

For large graphs, solving the MIP at a global level can become computationally demanding, as the number of decision variables and constraints grows exponentially with the dataset size. To improve performance, we add constraints only for instances $\vectorx$ and $\vectorx'$, such that $\vectorx' \in A_X$, reducing unnecessary computations. 
For full coverage, the complexity of the global Greedy approach is \( O(|X| k \log(d)) \) while the complexity for the local approach is \( O(m (k - m) |X_i| k \log(d)) \).

\section{FACEGroup for Auditing Fairness}
\label{seq: fairness}
In this section, we examine algorithmic fairness through the lens of FACEGroup. Group fairness refers to a set of principles designed to ensure that protected groups, often defined by sensitive attributes such as gender, race, or age, are treated similarly by a classifier. Broadly, group fairness can be categorized into \textit{demographic parity}, which requires that the proportion of positive outcomes reflects representation of the group in the population, and \textit{error-based fairness}, which focuses on equalizing classification errors, such as false negative rates, across groups~\cite{verma18,fragkathoulas2024explaining}.

To audit fairness for a group \(X\), we generate group counterfactual explanations (GCFs) for relevant subsets of \(X\). For example, we generate GCFs for the negatively classified instances of \(X\) when auditing for demographic parity, or the false negatives of \(X\) when auditing for error-based fairness. 
Disparities in the GCFs generated for different groups (e.g., males vs.\ females) can reveal potential biases in the model. 

Unlike existing approaches, FACEGroup supports \textit{multi-level} fairness auditing by partitioning each group into subgroups according to the connected components of the feasibility graph. 
This allows us to examine unfair behavior not only at the group level, but also at the level of subgroups, offering finer-grained insight into patterns of bias.
Furthermore, to capture the \textit{key trade-offs} in generating counterfactuals, FACEGroup provides novel fairness metrics that are parameterized by the number $k$ of counterfactuals, the cost $d$, and the coverage $c$. Introducing the number $k$ in the fairness metrics allows for assessing interpretability, as groups requiring fewer CFs are more interpretable, it promotes trust, as models that require fewer CFs are more transparent, and it serves in detecting disparities in CF requirements across (sub)groups, factors previously overlooked.

\textbf{Burden-based Fairness Measures.}
Counterfactuals provide a novel approach to measuring unfairness by evaluating both the disparities in outcomes between groups and the effort required by these groups to achieve fairness, i.e., to obtain the positive outcome. This effort, also called \textit{burden}, is often estimated as the aggregated cost between the factuals in a group and their counterfactuals \cite{10.1145/3375627.3375812,kuratomi2022measuring}. 
However, measuring burden solely at the group level may obscure disparities within subgroups, as different subpopulations may face varying degrees of difficulty in achieving favorable outcomes. 

We first define the 
minimum $k$ ($k_0$) and cost ($d_0$) required for full coverage (\( c = 1 \)):  
\begin{align*}
    k_0 &= \min \{k \mid \exists S, |S| \leq k, \text{coverage}(X, S) = |X| \}, \\
    d_0 &= \min \{d \mid \exists S, \text{cost}(X,S) \leq d, \text{coverage}(X, S) = |X| \}.
\end{align*}  
Note that \( k_0 \) is lower-bounded by the number of weakly connected components (\( k_0 \geq m \)), and \( d_0 \) does not exceed the largest WCC diameter.

We now introduce \textit{AUC-based fairness measures} that assess trade-offs between cost, number of counterfactuals, and coverage of (sub)groups across a range of parameter values rather at fixed points, avoiding biases from rigid parameter settings. 
The corresponding \textit{saturation points} identify optimal thresholds for cost, number of counterfactuals, and coverage.

We define the set of counterfactuals \( S_{k,d} \) that maximize coverage under a cost constraint \( d \) as:
\begin{align*}
    S_{k, d} &= arg max_{|S| \leq k, \, cost(X, S) \leq d} |coverage(X, S)|
\end{align*}
and $kAUC(k)$ as:
\begin{align*}
    kAUC(k) &= \int_{d_{min}}^{d_{max}} \text{coverage}(X, S_{k,d}) \, d{d}
\end{align*}  
that measures how efficiently a group can achieve coverage across a range of cost values for a given number of counterfactuals. 

\begin{figure}[ht]
    \centering
    \includegraphics[width=1\linewidth]{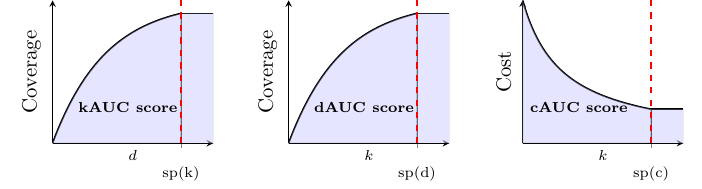}
    \caption{AUC scores and saturation points}
    \label{fig: auc_representation}
\end{figure}

Similarly, we define $dAUC(d)$ to evaluate how coverage improves as the number of counterfactuals increases under a fixed cost constraint, and $cAUC(c)$ to quantify the effort required to reach a given coverage level by measuring the total cost over a range of counterfactual numbers. 
Figure~\ref{fig: auc_representation} provides a visual representation of the AUC-based metrics.

There is also a minimum cost that provides the highest attainable coverage for $k$, we call it \textit{saturation point} for $k$ and denote it as $sp(k)$. Formally, it holds, for any $d \geq sp(k)$,  $coverage(X, S_{k, d}) = coverage(X, S_{k, sp(k)})$.
Similarly, we define,
$sp(d)$ to determine the least number of counterfactuals needed to reach maximum coverage within a given cost constraint, and 
$sp(c)$ to represent the minimum cost needed to achieve a desired coverage level, helping quantify the burden on different groups. Saturation points are shown in Figure~\ref{fig: auc_representation}.

\textbf{Attribution Measures.}
FACEGroup also provides insights into feature importance by measuring how often a feature change is required to alter an outcome. Concretely, the \textit{attribute change frequency} (\( ACF \)) metric captures how frequently a feature $A$ changes between a factual instance \( \vectorx \in X \) and its corresponding counterfactual \( \vectorx' \in S \):
\[
ACF(X, S, A) = \frac{1}{|X|} \sum_{\vectorx \in X} (1 - \delta(\vectorx_{A}, \vectorx_{A}')),
\]
where \( \delta(\vectorx_{A}, \vectorx_{A}') \) is the Kronecker delta, returning 1 if the feature remains unchanged and 0 otherwise. and \( \vectorx_{A} \) and \( \vectorx_{A}' \) represent the values of  \( A \) in the factual and counterfactual instances, respectively.
For each factual instance, we get the corresponding counterfactual instance with the minimum cost, i.e., $\vectorx' = argmin_{\vectorx'' \in S}cost(\vectorx, \vectorx'')$.

\section{Experimental Evaluation}
\label{seq: experiments}
The goal of our experimental evaluation is twofold: (a) to demonstrate the effectiveness of FACEGroup in fairness auditing and (b) to compare FACEGroup with baseline group counterfactual methods.

For fairness auditing, we use the widely studied \texttt{Adult}\footnote{\href{https://archive.ics.uci.edu/dataset/2/adult}{Adult}} dataset for income classification. To benchmark FACEGroup with baselines, we extend evaluations to additional datasets derived from US Census surveys, \texttt{AdultCA}\footnote{\href{https://github.com/socialfoundations/folktables}{Adult-CA-LA Datasets}}, \texttt{AdultLA}\footnotemark[2], and other domains including \texttt{COMPAS}\footnote{\href{https://www.propublica.org/datastore/dataset/compas-recidivism-risk-score-data-and-analysis}{COMPAS}}, \texttt{Student}\footnote{\href{https://archive.ics.uci.edu/dataset/297/student+performance}{Student}}, \texttt{German Credit}\footnote{\href{https://archive.ics.uci.edu/dataset/144/statlog+german+credit+data}{German Credit}}, and \texttt{HELOC}\footnote{\href{https://www.kaggle.com/datasets/averkiyoliabev/home-equity-line-of-creditheloc}{HELOC}}.
Further details on preprocessing, parameter settings, and configurations, as well as additional experiments on other datasets, are in the supplementary material. The source code is available online\footnote{\href{https://github.com/xristosfrag/FACEGroup-Feasible-and-Actionable-Counterfactual-Explanations-for-Group-Fairness-Auditing}{Project Repository}}.

First, we construct the feasibility graph $G_U$. An edge exists from a $\vectorxi$ to a $\vectorxj$ if the transition from  $\vectorxi$ to $\vectorxj$ is feasible and within threshold $\epsilon$.
We use a small set of generic feasibility constraints prohibiting unrealistic modifications, such as changing the values of immutable attributes (e.g., race) or the directionality of others, such as decreasing the value of the age attribute. 
The full set of constraints used is in the supplementary material.
We define groups based on the sensitive attribute \textit{Gender}: $G_0$ (females) and $G_1$ (males). 

\begin{figure}[ht]
    \centering
\includegraphics[width=10cm, height=2.5cm]{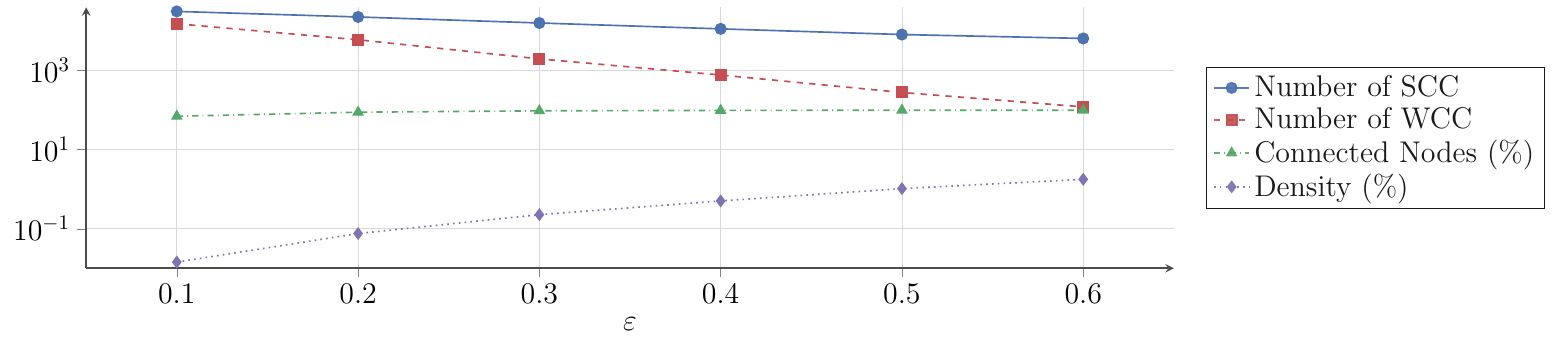}
    \caption{Feasibility graph connectivity based on the $\epsilon$ constraint.}
    \label{fig: graph_connectivity}
\end{figure}

Figure~\ref{fig: graph_connectivity} depicts the impact of varying $\epsilon$ on graph connectivity metrics, showing values up to the point where nearly all instances are connected, minimizing singleton nodes. Smaller $\epsilon$ values result in sparser graphs, ensuring that connected instances are more similar, leading to more plausible, small-step transitions. Conversely, larger $\epsilon$ values create denser graphs by incorporating connections between more distant instances, allowing for larger transition steps. 
To balance plausibility with connectivity, we select the smallest possible $\epsilon$ that maintains a highly connected graph while minimizing singleton nodes. For the \texttt{Adult} dataset, we set $\epsilon = 0.4$.
Further results for the selection of $\epsilon$ on the remaining datasets can be found in the supplementary material.

\subsection{Auditing Fairness}
In this set of experiments, we apply our algorithms to audit fairness. Without loss of generality, we focus on finding GCFs for the negatives for both groups $G_0$ and $G_1$. We use an XGBoost classifier optimized via hyperparameter tuning.
We consider only the instances in $G_0$ and $G_1$ for which at least one feasible candidate CF exists and use the $L_2$ distance as the cost function. 

\begin{figure}[t]
    \centering    
    \begin{subfigure}{0.48\textwidth}
        \centering
        \includegraphics[width=\linewidth]{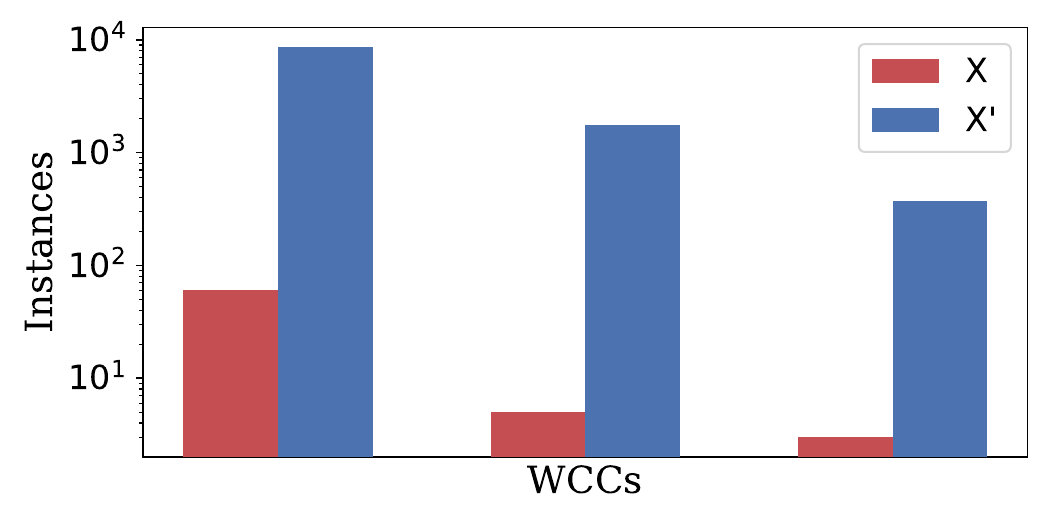} 
        \subcaption{$G_0$}
    \end{subfigure}
    \hfill
    \begin{subfigure}{0.48\textwidth}
        \centering
        \includegraphics[width=\linewidth]{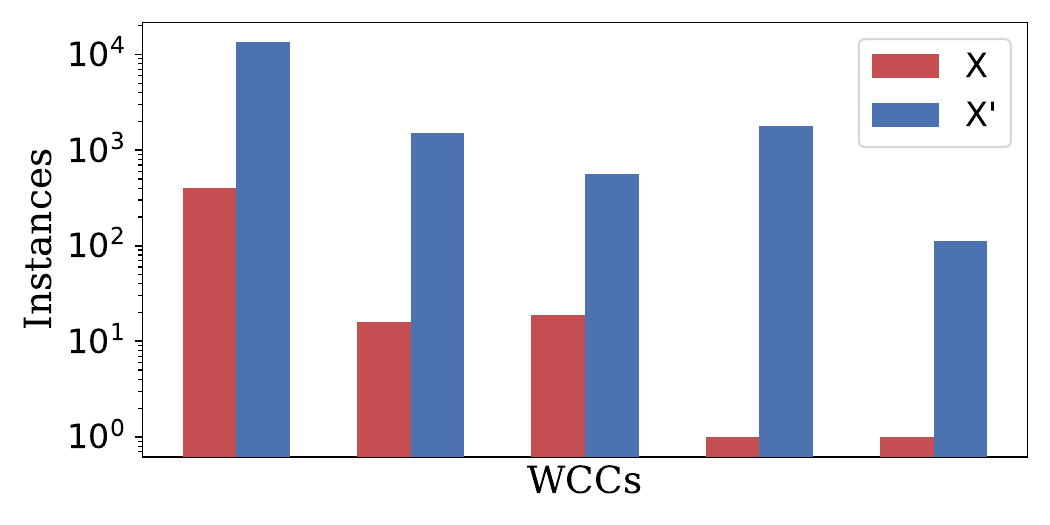} 
        \subcaption{$G_1$}
    \end{subfigure}
    \caption{Distribution of $\mathbf{X}$, $\mathbf{X'}$ per $WCC$ of the subgroups $G_0$ and $G_1$.}
    \label{fig: ccs_main_paper}
\end{figure}

\emph{\textbf{Burden Analysis.}} 
A key strength of FACEGroup is its ability to uncover subgroup behaviors within the groups $G_0$ and $G_1$ through the feasibility graph \( G_U \), which naturally partitions each group into WCCs, representing subpopulations that share feasible CF transformations. 
Figure~\ref{fig: ccs_main_paper} visualizes the distribution of factual instances (\textit{X}, red) and feasible counterfactual candidates (\textit{X'}, blue) across the subgroups (WCCs) of each group. We observe that $G_1$ exhibits a more fragmented structure, with CFs more widely spread across subgroups compared to $G_0$, suggesting that $G_1$ has a higher degree of variability in the transformations required for favorable outcomes.
Table \ref{tab: minimum_resources} depicts the minimum resources ($k_0$ and $d_0$) needed for full coverage per subgroup (WCC).
$G_1$ requires more CFs ($k_0 = 12$) than $G_0$ ($k_0 = 9$) and higher minimum cost ($d_0 = 1.04$) than $G_0$ ($d_0 = 0.93$), suggesting greater heterogeneity in the CF pathways needed for full coverage.

\begin{table}[ht]
  \centering
  \captionsetup{justification=centering}
  \caption{$k_0$ and $d_0$ for each $WCC$ of each group and overall for each group.}
  \label{tab:combined}
  \newcolumntype{C}[1]{>{\centering\arraybackslash}p{#1}}
  \begin{tabular}{
    C{0.8cm}|
    *{5}{C{0.8cm}C{0.8cm}}|
    C{0.9cm}C{0.9cm}
  }
    \toprule
    \multicolumn{11}{c}{WCCs} & \multicolumn{2}{c}{Overall} \\
    \cmidrule(lr){2-11}\cmidrule(lr){12-13}
    \multirow{2}{*}{} &
    \multicolumn{2}{c}{$WCC_1$} & 
    \multicolumn{2}{c}{$WCC_2$} &
    \multicolumn{2}{c}{$WCC_3$} & 
    \multicolumn{2}{c}{$WCC_4$} & 
    \multicolumn{2}{c|}{$WCC_5$} & 
    \textbf{$k_0$} & \textbf{$d_0$} \\
    \cmidrule(lr){2-11} \cmidrule(lr){12-13}
    & $k_0$ & $d_0$ & $k_0$ & $d_0$ & $k_0$ & $d_0$ & $k_0$ & $d_0$ & $k_0$ & $d_0$ & & \\
    \midrule
    $G_0$ & 7 & 0.93 & 1 & 0.74 & 1 & 0.49 & -- & -- & -- & -- & 9 & 0.93 \\
    $G_1$ & 4 & 1.04 & 3 & 0.61 & 3 & 0.78 & 1 & 0.46 & 1 & 0.20 & 12 & 1.04 \\
    \bottomrule
  \end{tabular}
  \label{tab: minimum_resources}
\end{table}

Analyzing subgroups is crucial, as group-level fairness assessments can mask heavily disadvantaged subpopulations, leading to misleading conclusions about the equitable distribution of the burden.
At the subgroup level, the \textit{Black} subgroups (that correspond to  $WCC_1$ in both groups) exhibit the highest $k_0$ and $d_0$, indicating that
they face greater barriers to obtain favorable decisions. Notably, the subgroups with the most factual instances also bear the highest burden, indicating a disproportionate impact on overall group difficulty.

Table \ref{tab: combined_auc_adult} reports $kAUC$, $dAUC$, $cAUC$, saturation points $sp$, and the minimum, or maximum values for coverage and cost, that correspond to each $sp$. Scores are normalized by the optimal $AUC$ per metric. Higher $kAUC$, $dAUC$ and lower $cAUC$ are preferred. 
 
 \begin{table}[h]
    \centering
    \setlength{\tabcolsep}{6pt}    
    \caption{$kAUC$, $dAUC$, $cAUC$, and saturation points.}
    \label{tab: combined_auc_adult}
    \begin{tabular}{c c|c c c|c c c}
        \hline
        \textbf{Parameter} & \textbf{Value} & \multicolumn{3}{c|}{\(G_0\)} & \multicolumn{3}{c}{\(G_1\)} \\
        \cline{3-8}
        \hline
        \multicolumn{8}{c}{\ensuremath{\textbf{\textit{kAUC}}} metrics} \\
        \hline
        &  & \(sp(k)\) & Max Cov. & \(kAUC\) & \(sp(k)\) & Max Cov. & \(kAUC\) \\
        \cline{3-8}
         \multirow{4}{*}{$k$} & 1  & 1.1  & 63.08 & 0.50  & 1.3  & 65.75 & 0.54 \\
         & 5  & 1.1  & 93.85 & 0.82  & 1.1  & 97.49 & 0.85 \\
         & 9  & 1.1  & 100.0 & 0.90  & 1.1  & 99.09 & 0.89 \\
         & 13 & 0.7  & 100.0 & 0.92  & 1.1  & 100.0 & 0.91 \\
        \hline
        \multicolumn{8}{c}{\ensuremath{\textbf{\textit{dAUC}}} metrics} \\
        \hline
        &  & \(sp(d)\) & Max Cov. & \(dAUC\) & \(sp(d)\) & Max Cov. & \(dAUC\) \\
        \cline{3-8}
         \multirow{4}{*}{$d$} & 0.1 & 6  & 12.31 & 0.10  & 12  & 12.78 & 0.08 \\
         & 0.8 & 10  & 100.0 & 0.89  & 12  & 99.31 & 0.93 \\
         & 1.5 & 9  & 100.0 & 0.93  & 12  & 99.77 & 0.95 \\
         & 2.2 & 9  & 100.0 & 0.93  & 12  & 99.77 & 0.95 \\
        \hline
        \multicolumn{8}{c}{\ensuremath{\textbf{\textit{cAUC}}} metrics} \\
        \hline
        &  & \(sp(c)\) & Min Cost & \(cAUC\) & \(sp(c)\) & Min Cost & \(cAUC\) \\
        \cline{3-8}
        \multirow{4}{*}{$c$} & 0.25 & 12 & 0.14 & 0.10  & 20 & 0.12 & 0.11 \\
                             & 0.50 & 18 & 0.22 & 0.17  & 23 & 0.20 & 0.17 \\
                             & 0.75 & 22 & 0.28 & 0.25  & 25 & 0.30 & 0.25 \\
                             & 1.00 & 16 & 0.55 & 0.56  & 20 & 1.40 & 0.72 \\
        \hline
    \end{tabular}
    \end{table}

For $kAUC$, saturation points ($sp$) are expected to decrease as more CFs are provided. Initially, at $k=1$, $G_1$ achieves higher maximum coverage, reflecting larger available transitioning costs, enabling more instances to be efficiently covered at low $k$. However, as the number of CFs increases, 
$G_0$ reaches full coverage first, exhibiting better overall efficiency (higher $kAUC$) and requiring fewer resources (lower $sp$ values) compared to $G_1$.
For $dAUC$, saturation points should decrease as higher-cost connections are allowed.
At $d=0.1$, $G_0$ has a lower $sp(d)$, indicating fewer feasible low-cost available transitions, compared to $G_1$. 
As cost increases, $G_0$ effectively utilizes connections to reach full coverage with fewer CFs, while $G_1$ requires higher costs to achieve maximum comparable coverage.
However, when
$d \in [0.8, 1.5]$, $G_1$ exhibits stronger coverage efficiency gains, suggesting $G_0$ is more efficient at lower costs while $G_1$ benefits more from cost relaxations.
For $cAUC$, both groups experience similar cost burdens for achieving intermediate coverage levels \(0.25, 0.5\) and \(0.75\). However, at full coverage ($c=1.0$), $G_1$ incurs significantly higher costs, as reflected in both $cAUC$ and minimum cost. The consistently higher $sp(c)$ values for $G_1$ suggest that more CFs are required to reach cost-efficient solutions, reinforcing a systemic disadvantage in obtaining full coverage at minimal cost while maintaining interpretability.

\begin{figure*}[t]
    \centering   
    \begin{subfigure}{0.32\textwidth}
        \centering
        \includegraphics[width=\linewidth]{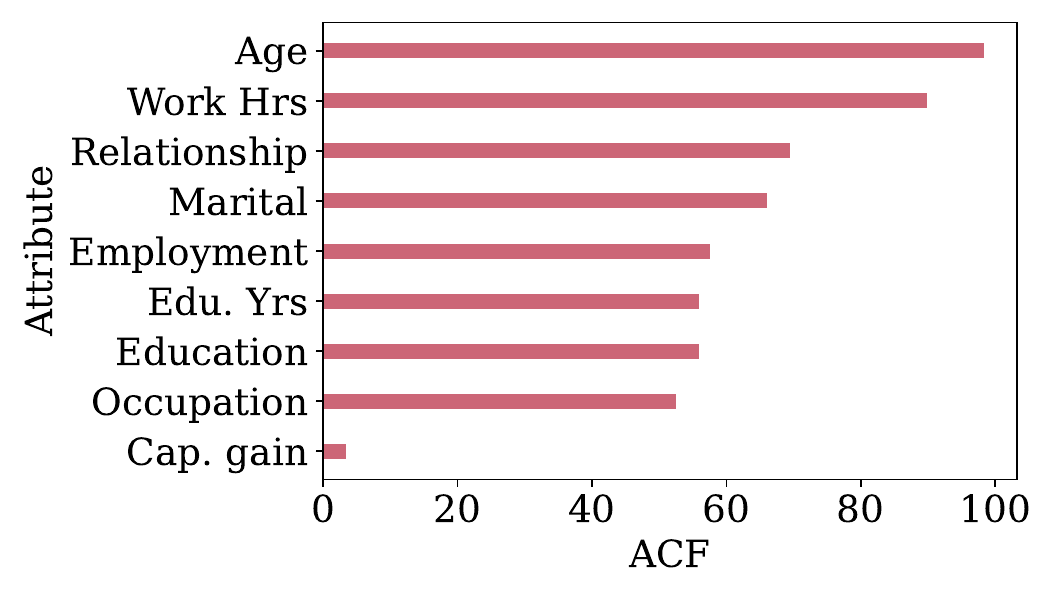} 
        \caption{$G_0: WCC 1$}
    \end{subfigure}
    \hfill
    \begin{subfigure}{0.32\textwidth}
        \centering
        \includegraphics[width=\linewidth]{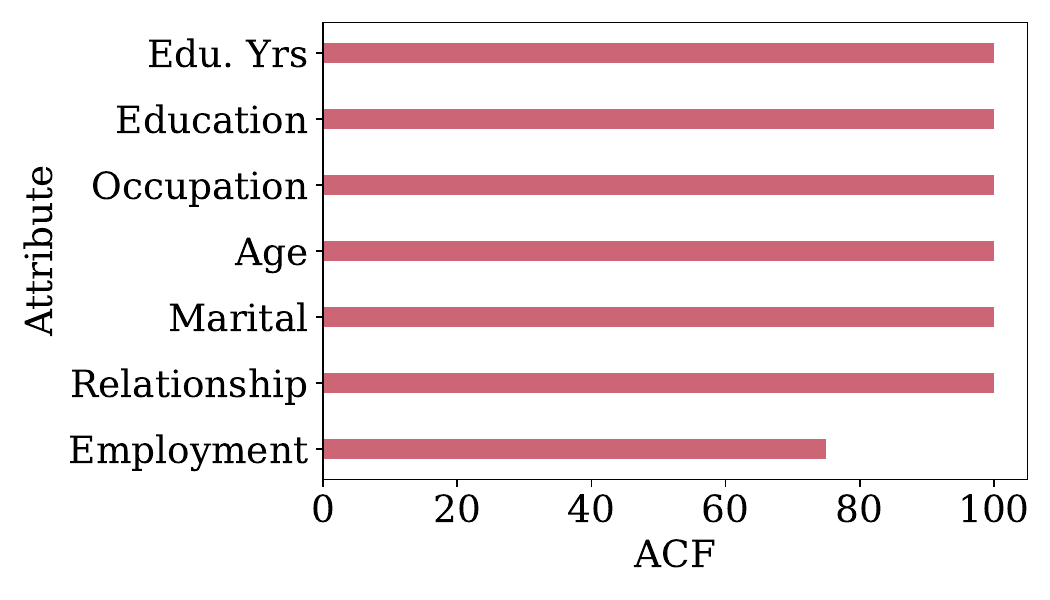} 
        \caption{$G_0: WCC 2$}
    \end{subfigure}
    \hfill
    \begin{subfigure}{0.32\textwidth}
        \centering
        \includegraphics[width=\linewidth]{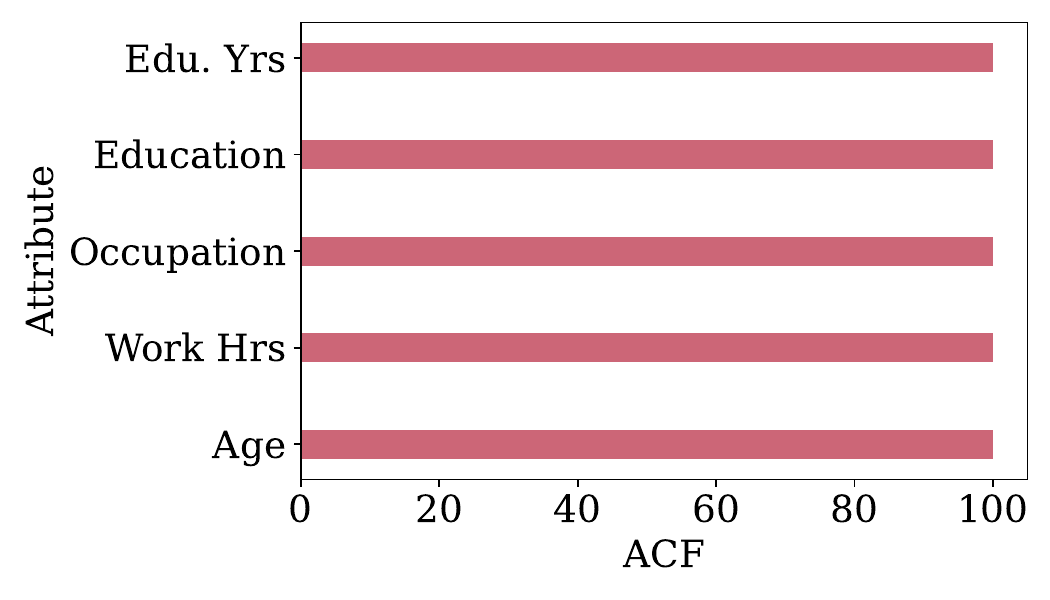} 
        \caption{$G_0: WCC 3$}
    \end{subfigure}
    \begin{subfigure}{0.32\textwidth}
        \centering
        \includegraphics[width=\linewidth]{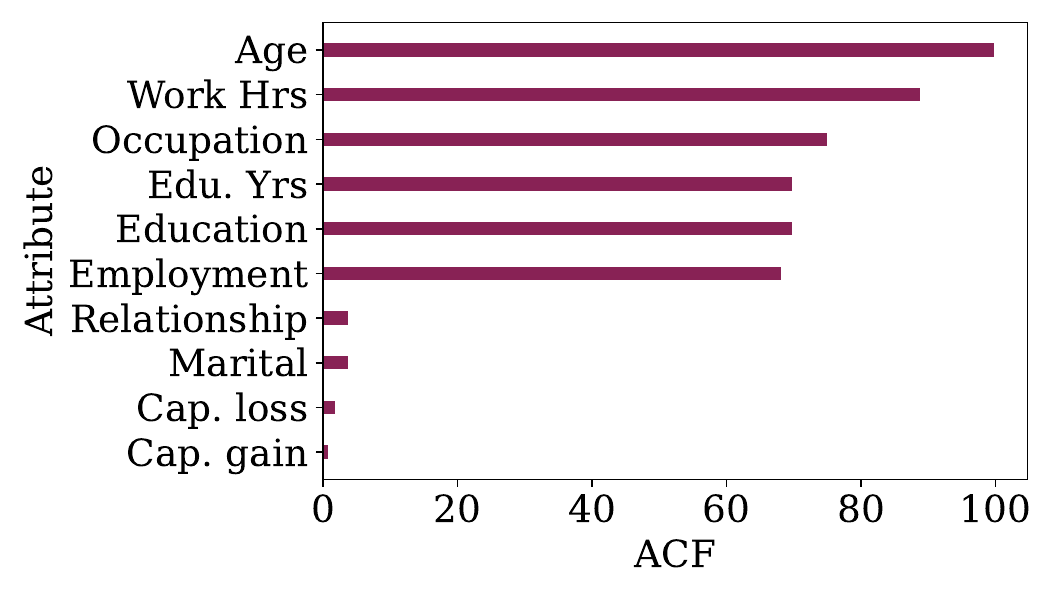} 
        \caption{$G_1: WCC 1$}
    \end{subfigure}
    \hfill
    \begin{subfigure}{0.32\textwidth}
        \centering
        \includegraphics[width=\linewidth]{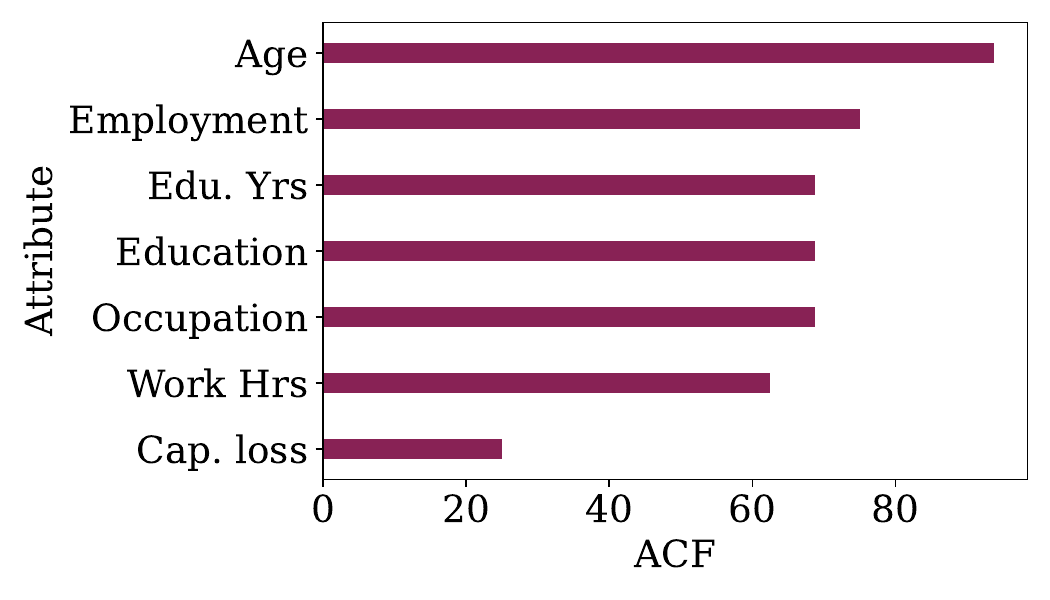}
        \caption{$G_1: WCC 2$}
    \end{subfigure}
    \hfill
    \begin{subfigure}{0.32\textwidth}
        \centering
        \includegraphics[width=\linewidth]{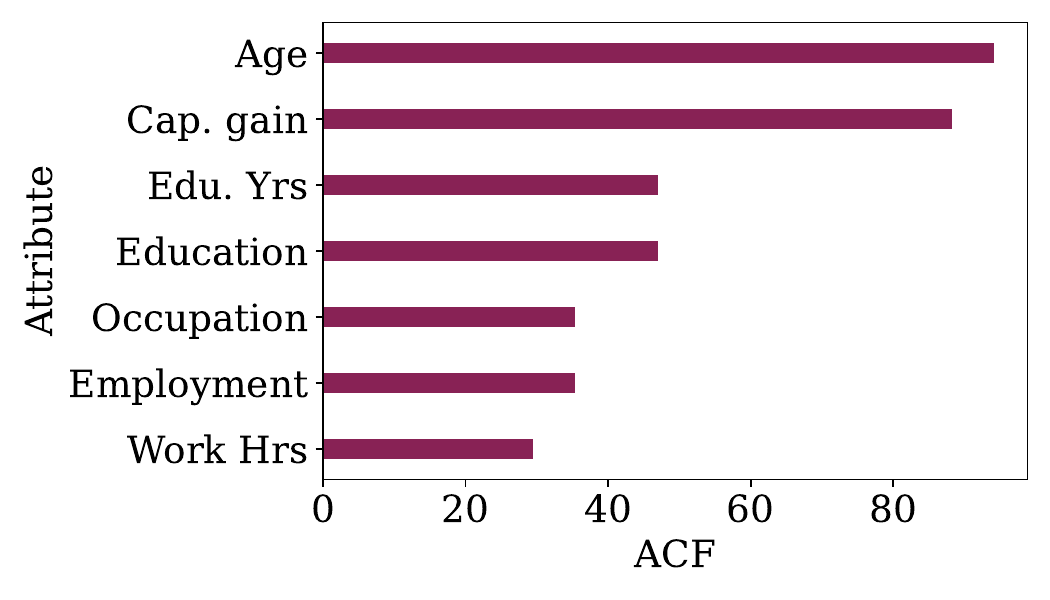} 
        \caption{$G_1: WCC 3$}
    \end{subfigure}
    \caption{ACF across the subgroups of each group}
    \label{fig: attribution}
\end{figure*}

\emph{\textbf{Attribution Analysis.}} 
To further analyze subgroup disparities, we use the \textit{ACF} metric per WCC, quantifying how often specific attributes are altered in CFs, providing insights into the different factors driving classification decisions.
Figure~\ref{fig: attribution} presents the frequency of modified attributes for each WCC of \( G_0 \) and \( G_1 \), respectively, and shows that subgroup-specific variations exist in the importance of different attributes. For \( G_1 \), we include only the three largest WCCs, excluding those with few factual instances, as they lack representativeness.
A common trend across all WCCs in both groups is that an increase in \textit{age} is frequently required for a favorable outcome, suggesting that the model associates age with work experience or financial stability.  
Within $G_0$, the Asian-Pacific-Islander individuals ($WCC_3$) require fewer modifications compared to the Blacks ($WCC_1$) and Amer-Indian-Eskimos ($WCC_2$) and do not rely on \textit{relationship status} or \textit{marital status}, unlike the others.  
In $G_1$, despite similar CF difficulty (Table \ref{tab: minimum_resources}), financial interventions differ: Amer-Indian-Eskimos ($WCC_2$) require career-related changes (\textit{employment status, occupation, education}), while Asian-Pac-Islanders ($WCC_3$) depend on increasing \textit{capital gain}.
More broadly, \textit{capital gain} is largely absent from both groups of CFs except for $G_1 - WCC_3$, highlighting subgroup differences in financials to favorable outcomes. 
Finally, CFs in $G_1$ rarely modify \textit{relationship status}, unlike in $G_0$, where it is frequently altered. Instead, \textit{educational} and \textit{occupational} factors are highly important.

\subsection{Comparison with Baselines}
We evaluate FACEGroup against existing CF generation methods, specifically: (a) with FACE \cite{poyiadzi2020face}, a graph-based method for individual CFs, and (b) with AReS \cite{rawal2020beyond} and GLOBE-CE \cite{ley2023globe}, two state-of-the-art GCF approaches. 

\textbf{Comparison with Individual CFs}
Given a group $X$, FACEGroup generates a small set $S$ of $k$ counterfactuals to cover $X$. To evaluate the efficiency of this approach, we compare the associated cost with the cost of generating \textit{individual counterfactuals} for each instance in $X$, which serves as a lower bound on the cost when the constraint on $k$ is relaxed.
For generating individual counterfactuals, we use FACE, since it is also based on a feasibility graph. 
For these experiments, we generate CFs for the full population $G = G_0 \cup G_1$.
We assess how closely GCFs from FACEGroup approximate the optimal costs of individual CFs from FACE.
First, we apply FACEGroup to generate the set $S$ of CFs by solving the coverage-constrained problem.
Then, we apply FACE to all factuals covered by $S$
using the same cost function. As a cost function, we use both: (a) the weighted shortest path cost in $G_U$ (originally used in FACE), and (b) the  $L_2$ distance.

\begin{figure}[ht]
    \centering
    \begin{subfigure}{.325\textwidth}
        \centering
        \includegraphics[width=\linewidth]{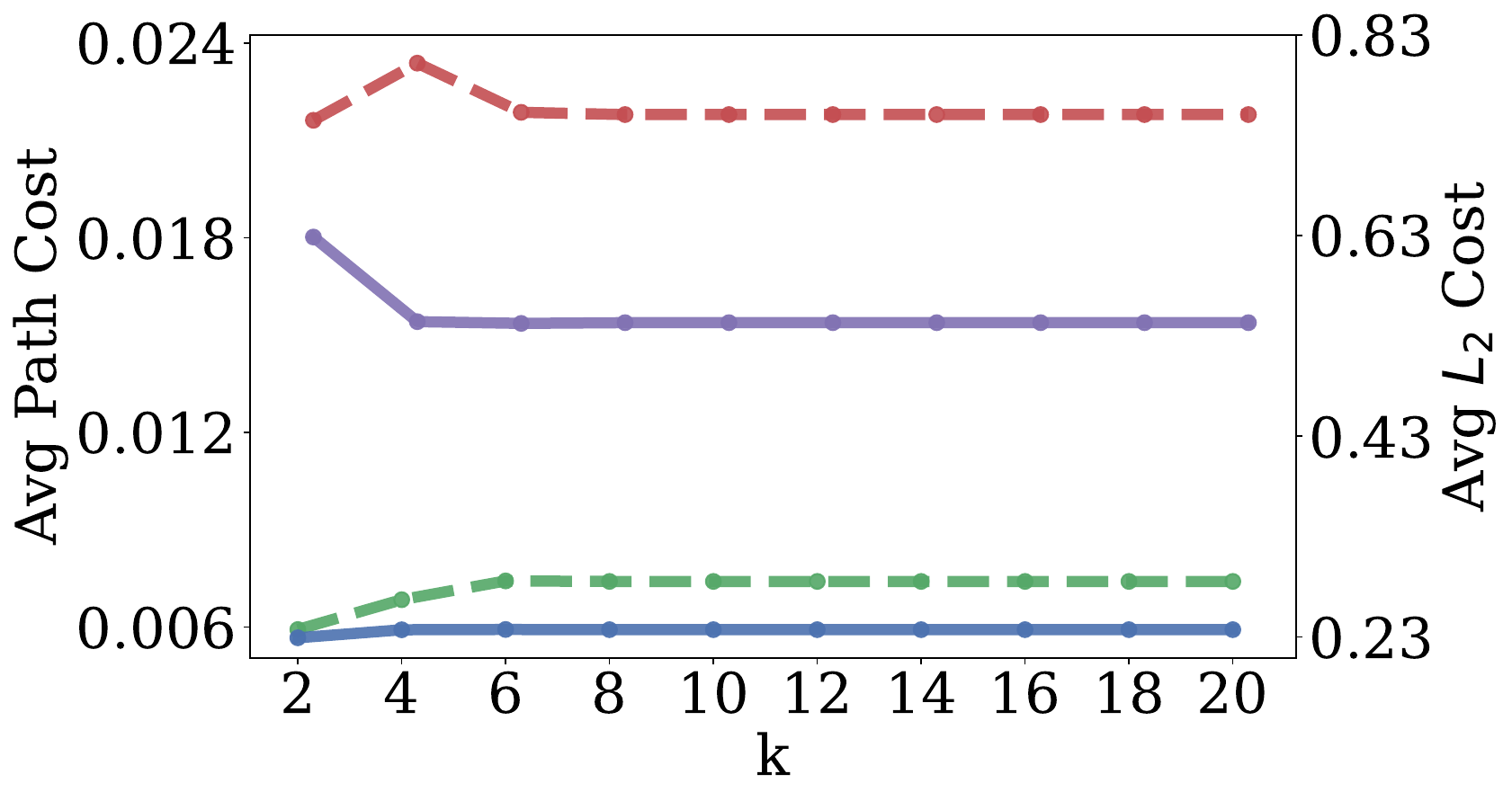}
    \end{subfigure}
    \hfill
    \begin{subfigure}{.325\textwidth}
        \centering
        \includegraphics[width=\linewidth]{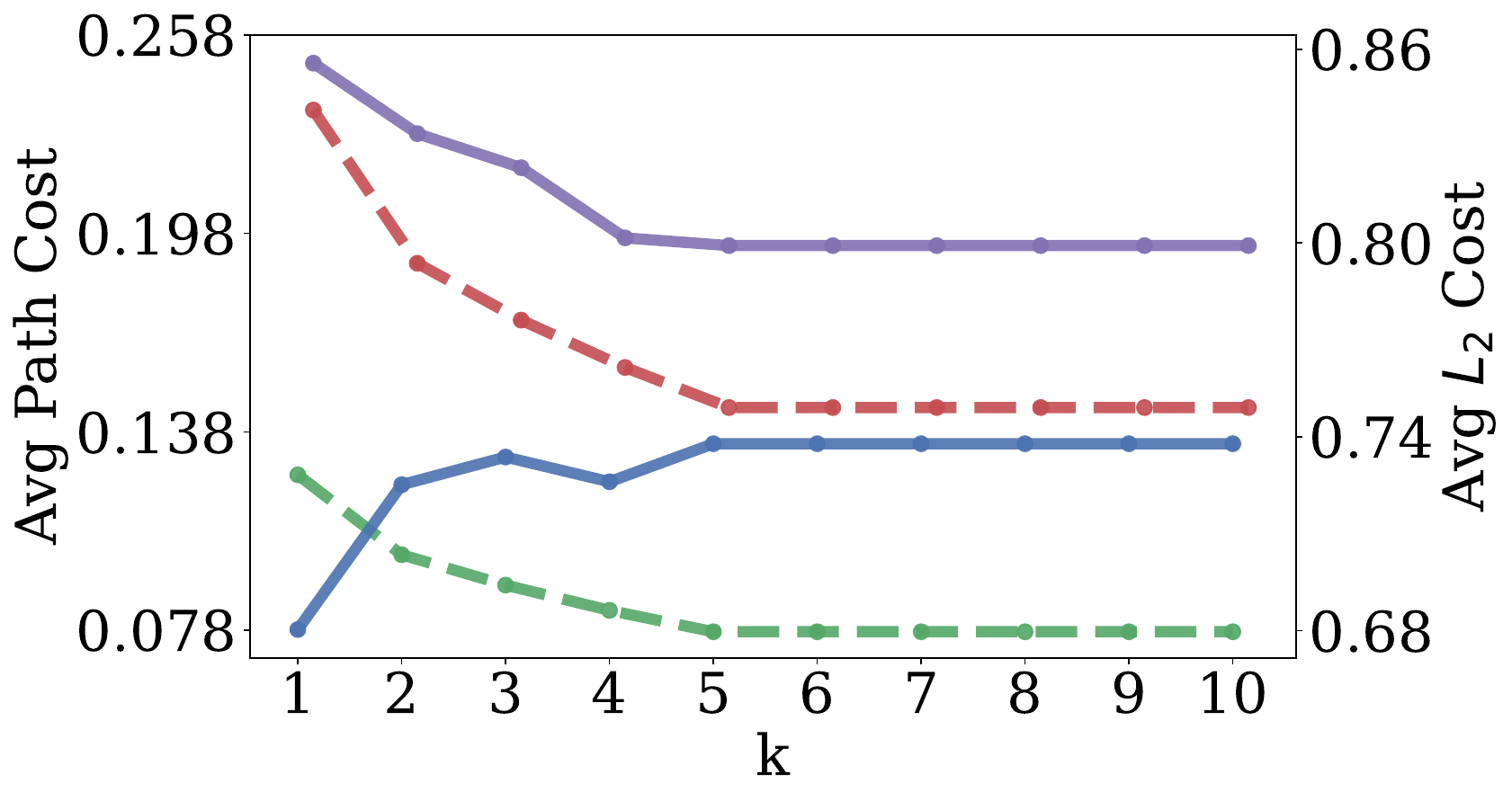}
    \end{subfigure}
    \hfill
    \begin{subfigure}{.325\textwidth}
        \centering
        \includegraphics[width=\linewidth]{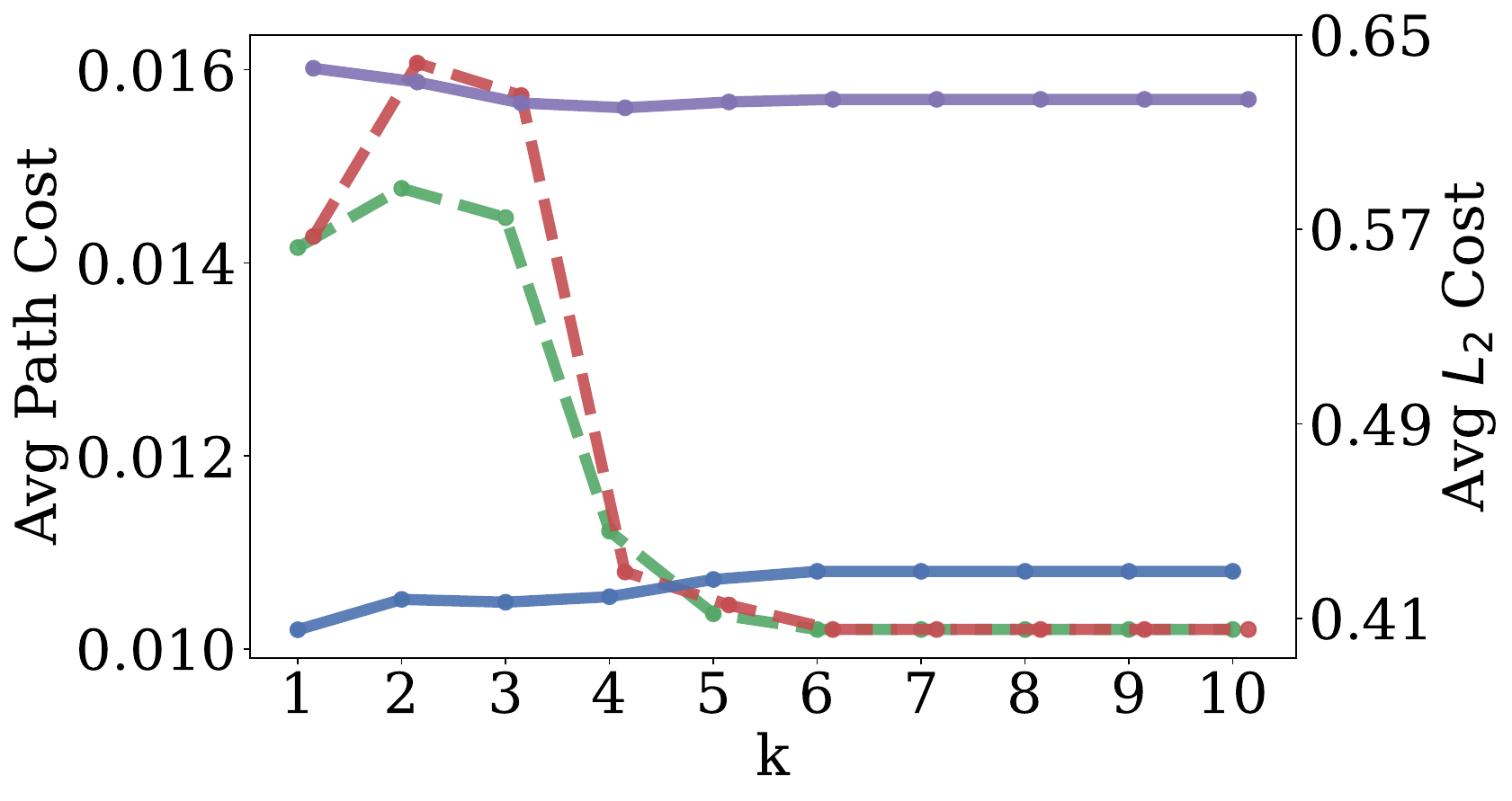}
    \end{subfigure}

    \includegraphics[width=0.95\textwidth]{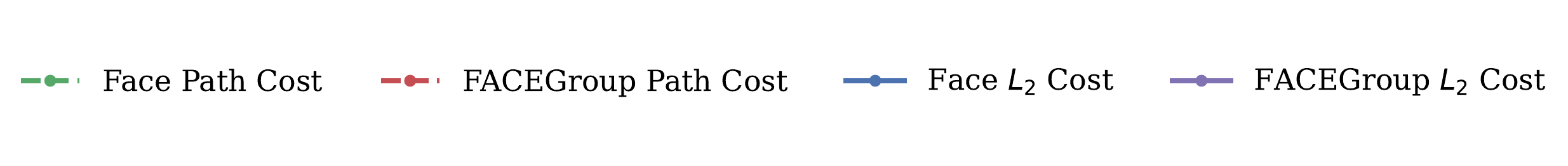}

    \begin{subfigure}{.325\textwidth}
        \centering
        \subcaption{\texttt{Adult} Dataset}
    \end{subfigure}
    \hfill
    \begin{subfigure}{.325\textwidth}
        \centering
        \subcaption{\texttt{German Credit} Dataset}
    \end{subfigure}
    \hfill
    \begin{subfigure}{.325\textwidth}
        \centering
        \vspace{-10em}
        \subcaption{\texttt{HELOC} Dataset}
    \end{subfigure}
    
    \captionsetup{justification=centering}
    
    \caption{Comparison of FACEGroup and FACE on average CF costs.}
    \label{fig: face_main_paper}
\end{figure}

Figure~\ref{fig: face_main_paper} shows the cost comparison for $k$ CFs from 1 to $k_0$ in 10 equal steps, with normalized costs. As expected, FACE achieves the lowest costs, while FACEGroup, which prioritizes group-level explanations, incurs slightly higher but still near-optimal costs. FACEGroup maintains near-optimal shortest path costs in datasets like \texttt{German Credit} and \texttt{HELOC}, where feasible transformations remain efficient. However, in \texttt{Adult}, costs increase due to the challenge of balancing feasibility with compact group CFs. Similar trends hold across other datasets, with full results and parameter details provided in the supplementary material.

\textbf{Comparison with GCF Methods}  
We compare FACEGroup with two state-of-the-art GCF baselines: AReS \cite{rawal2020beyond} and GLOBE-CE \cite{ley2023globe}. AReS mines frequent itemsets from individuals who achieved the desired outcome, selecting a small, interpretable set of rules via a submodular objective. GLOBE-CE defines global CFs as translation vectors applied to groups, scaling them across a range of values to adapt to individuals.

Both baselines without feasibility and plausibility constraints achieve at least $70\%$ coverage.
AReS generates 3 to 20 rules, while GLOBE-CE produces a significantly larger set, ranging from 10 to 612 CFs, due to the multiple scales on top of the translation vectors. Detailed results are in the supplementary material.
To assess feasibility, we integrate CFs into the feasibility graph $G_U$ and measure feasibility coverage as the proportion of CFs with at least one feasible transition. We analyze this under all feasibility constraints and a relaxed setting with only the plausibility constraint $\epsilon$. Table~\ref{tab:baselines} highlights the limitations of baselines: with full constraints, AReS and GLOBE-CE remain below 50\% feasibility coverage, indicating that many CFs violate real-world constraints. In contrast, FACEGroup achieves 100\% feasibility coverage with a compact CF set.
Relaxing constraints improves coverage for baselines, particularly for GLOBE-CE, which benefits from its low-cost translation vectors. 
However, FACEGroup still maintains full feasibility coverage with fewer CFs, demonstrating its ability to generate feasible, actionable CFs without sacrificing interpretability or plausibility.
\begin{table*}[t] 
\centering
\renewcommand{\arraystretch}{0.6}
\setlength{\tabcolsep}{6pt}
\caption{Comparison with baselines.}
\label{tab:baselines}
\begin{tabular}{lcccccccc}
\toprule
Dataset & $\epsilon$ & FC & \multicolumn{2}{c}{AReS} & \multicolumn{2}{c}{GLOBE-CE} & \multicolumn{2}{c}{FACEGroup}\\ 
\cmidrule(lr){4-5} \cmidrule(lr){6-7}\cmidrule(lr){8-9}
& & & $r$ & Cov. (\%) & $k$ & Cov. (\%) & $k$ & Cov. (\%)\\
\midrule
\multirow{2}{*}{Adult}
& 0.4 & all & 18 & 15.68 & 421 & 0.24 & 21& 100\\
& 0.4 & none & 18 & 52.26 & 421& 84.56 & 10 & 100\\
\midrule
\multirow{2}{*}{AdultCA}
& 0.7  & all &20 &11.36 & 612& 11 &133&100\\
& 0.7 & none & 20& 11.36& 612& 11.50 &15& 100\\
\midrule
\multirow{2}{*}{AdultLA}
& 0.5 & all & 20 & 12.9 & 342 & 12.9 &59&100\\
& 0.5 & none & 20 & 23.11 & 342 & 22.63 & 13 & 100\\

\midrule
\multirow{2}{*}{Student}
& 3.0 & all &3 & 33.3 & 10 & 50  & 3 & 100\\
& 3.0 & none & 3 & 75 & 10 & 66.67 & 2 & 100\\
\midrule
\multirow{2}{*}{COMPAS}
& 0.3 & all &20 & 11.85  & 124 &20  & 13 &100\\
& 0.3 & none & 20 & 16.3 &124  & 25.93 & 13 & 100\\
\midrule
\multirow{2}{*}{German Credit}
& 2.9 & all &4 & 0 & 18 &26.32  & 6 &100\\
& 2.9 & none & 4 & 42.11  & 18 & 73.68 & 2 & 100\\
\midrule
\multirow{2}{*}{HELOC}
& 1.4  & all & 11& 1.98 & 74 & 1 & 4 & 100\\
& 1.4 & none & 11 & 71.29 & 74 & 72.28 &2  & 100\\
\bottomrule
\end{tabular}
\end{table*}

\section{Related Work}
\label{seq: related}
Explanations have become central in machine learning research 
\cite{fragkathoulas2024explaining,guidotti2022counterfactual},
particularly in high-stakes domains such as healthcare and education. Among various explanation methods, CFs have gained prominence for their ability to reveal actionable changes leading to a desired outcome.
Wachter et al. \cite{wachter2017counterfactual} first formulated CFs as an optimization problem, minimizing the cost between an instance and its CF while ensuring a prediction change. Subsequent work \cite{guidotti2019factual,mothilal2020explaining,kanamori2020dace,goethals2024precof,10.1145/3677119,10.1145/3375627.3375812,poyiadzi2020face} refined CF generation, emphasizing properties such as feasibility, actionability, sparsity \cite{10.1145/3677119}, and robustness\cite{guyomard2023generating}.
Several approaches optimize CF search using genetic algorithms \cite{10.1145/3375627.3375812,fragkathoulas2025ugce}, integer programming \cite{russell2019efficient,ustun2019actionable}, and cost-based heuristics \cite{goethals2024precof}.

FACE \cite{poyiadzi2020face} constructs a density-weighted feasibility graph where counterfactuals are generated via shortest paths in the graph, focusing on individual explanations that balance proximity and data manifold alignment. While FACEGroup builds on this graph structure, and further introduces three key innovations: (1) multi-level subgroup analysis, where WCCs of the feasibility graph naturally partition groups into interpretable subgroups with shared feasibility constraints, (2) GCF trade-off-aware algorithms, rather than relying on individual shortest-path searches, and (3) cost function agnosticism.

While most methods focus on individual CFs, recent work explores GCFs for multiple instances. AReS \cite{rawal2020beyond} defines subgroup-specific CF rules, optimizing for correctness, coverage, cost, and interpretability. GLOBE-CE \cite{ley2023globe} learns global translation vectors, applying them at different scales to generate CFs that maximize coverage. CET \cite{kanamori2022counterfactual} uses decision trees for group actions to enhance transparency and consistency, while mixed-integer programming has been used to optimize collective CFs under linking constraints \cite{carrizosa2024generating}. CounterFair \cite{kuratomi2024counterfair} generates fair GCFs by selecting a subset via mixed-integer programming to balance cost and fairness.
Unlike these approaches, FACEGroup enforces feasibility constraints, ensuring GCFs adhere to real-world constraints.
Most group-based methods only prevent changes in sensitive attributes but lack directional constraints, leading to CFs that may violate plausible transformations. Notably, GLOBE-CE selects random feature perturbations, which can result in unrealistic CFs. In contrast to these methods, FACEGroup generates CFs at both group and subgroup levels, systematically handling the trade-offs in CF generation.

Explanations are utilized to assess algorithmic fairness \cite{fragkathoulas2024explaining}, ensuring decisions are not influenced by protected attributes \cite{pitoura2022fairness,mehrabi2021survey,friedler2021possibility,caton2024fairness}.
Several CF-based approaches have been proposed to quantify fairness by measuring the burden quantified as the difficulty individuals face in achieving a favorable outcome per group \cite{kuratomi2022measuring,10.1145/3375627.3375812,goethals2024precof,kavouras2024fairness,rawal2020beyond}. 
Methods like \cite{10.1145/3375627.3375812,kuratomi2022measuring} generate individual CFs and calculate burden per group as the average sum of pairwise costs to assess fairness.
PreCoF \cite{goethals2024precof} distinguishes between explicit bias, when individual counterfactuals require changes only in sensitive attributes, and implicit bias, when, after removing sensitive attributes from model training, other features disproportionately influence different groups.
\cite{rawal2020beyond,ley2023globe} suggest that generated rules and global translation vectors can be used to manually audit for unfairness in subgroups of interest.
FACTS \cite{kavouras2024fairness} builds on AReS and introduces burden-based fairness metrics, but evaluates fairness only under specific settings.
For instance, its Equal Cost of Effectiveness metric compares the minimum cost needed for protected subgroups to reach a fixed aggregate effectiveness level, defined as the proportion of individuals able to achieve the desired outcome via counterfactuals. In contrast, our burden-based fairness metrics assess disparities across a range of costs, coverage levels, and numbers of counterfactuals, offering a more comprehensive perspective that captures potential disparities across various combinations of these factors. 
Unlike the other approaches, FACEGroup introduces fairness metrics that assess fairness at both group and subgroup levels, explicitly accounting for trade-offs between cost, coverage, interpretability, and feasibility.

\section{Conclusions}
\label{seq: conclusion}
In this paper, we propose FACEGroup, a novel graph-based framework for group counterfactual generation that addresses limitations in existing methods by incorporating real-world feasibility constraints and managing trade-offs in counterfactual generation. 
We also introduce novel fairness measures that allow auditing fairness both at the group and subgroup levels, offering insights on the trade-offs between cost, the number of generated counterfactuals, and coverage.
In future work, we plan to extend the use of the feasibility graph to define path-based fairness metrics. We also aim to adapt our approach to multi-class classification and regression settings.

\section{Acknowledgment}
This work has been partially supported by project MIS 5154714 of the National Recovery and Resilience Plan Greece 2.0 funded by the European Union under the NextGenerationEU Program.

\appendix
\section{Datasets and Feasibility Graph Construction}
\label{seq: datasets_graph}
In this section, we describe the datasets, specify the feasibility constraints used for each, and discuss the selection of the $\epsilon$ parameter in constructing the feasibility graph.
\subsection{Dataset Descriptions}
\label{section: Supplementary Material}
We evaluate FACEGroup on six datasets. The \texttt{Adult}\footnote{\href{https://archive.ics.uci.edu/dataset/2/adult}{Adult Dataset}} dataset to showcase the effectiveness of FACEGroup to discover both group and subgroup disparities, and 5 more datasets that support the superiority of FACEGroup against baselines. 
Two more recent datasets derived from US Census surveys\cite{ding2021retiring}, ~\texttt{Adult- California}~\footnote{\href{https://github.com/socialfoundations/folktables}{Adult-California-Louisana Datasets}} (AdultCA) and \texttt{Adult-Louisiana}\footnotemark[12] (AdultLA), obtained from the \textit{ACS PUMS dataset}. For \texttt{AdultCA} and \texttt{AdultLA}, we select data from 2023, as it represents the most up-to-date information available. These three datasets consist of records of individuals used for predicting if their annual income exceeds $\$50,000$. 
The \texttt{Student}\footnote{\href{https://archive.ics.uci.edu/dataset/297/student+performance}{Student Dataset}}  dataset includes records of student performance, featuring attributes like study time and family support. The target labels are derived from the final grade (G3), with students categorized as having lower performance if G3 is less than $10$ and high performance if G3 is $10$ or higher.
\texttt{COMPAS}\footnote{\href{https://www.propublica.org/datastore/dataset/Compas-recidivism-risk-score-data-and-analysis}{COMPAS Dataset}}, contains instances from the criminal justice system used to predict the likelihood of recidivism. 
\texttt{German Credit}\footnote{\href{https://archive.ics.uci.edu/dataset/144/statlog+german+credit+data}{German Credit Dataset}} dataset classifies individuals as either good or bad credit risks based on various attributes such as credit history, account status, and employment and the \texttt{HELOC}\footnote{\href{https://www.kaggle.com/datasets/averkiyoliabev/home-equity-line-of-creditheloc}{HELOC Dataset}}, consists of credit card account information, including attributes such as credit limits, payment history, and credit utilization rates, used for predicting credit risk or the likelihood of default. 
The numerical attributes where the values represent measurements or quantities with many unique values are treated as continuous.
More details about each attribute description, feasibility constraint, and corresponding datatype of \texttt{Adult, AdultCA, AdultLA, Student, COMPAS, German Credit} and \texttt{HELOC} can be found in Table \ref{dataset: datasets}.

\begin{figure}
    \centering
    \includegraphics[width=1\linewidth]{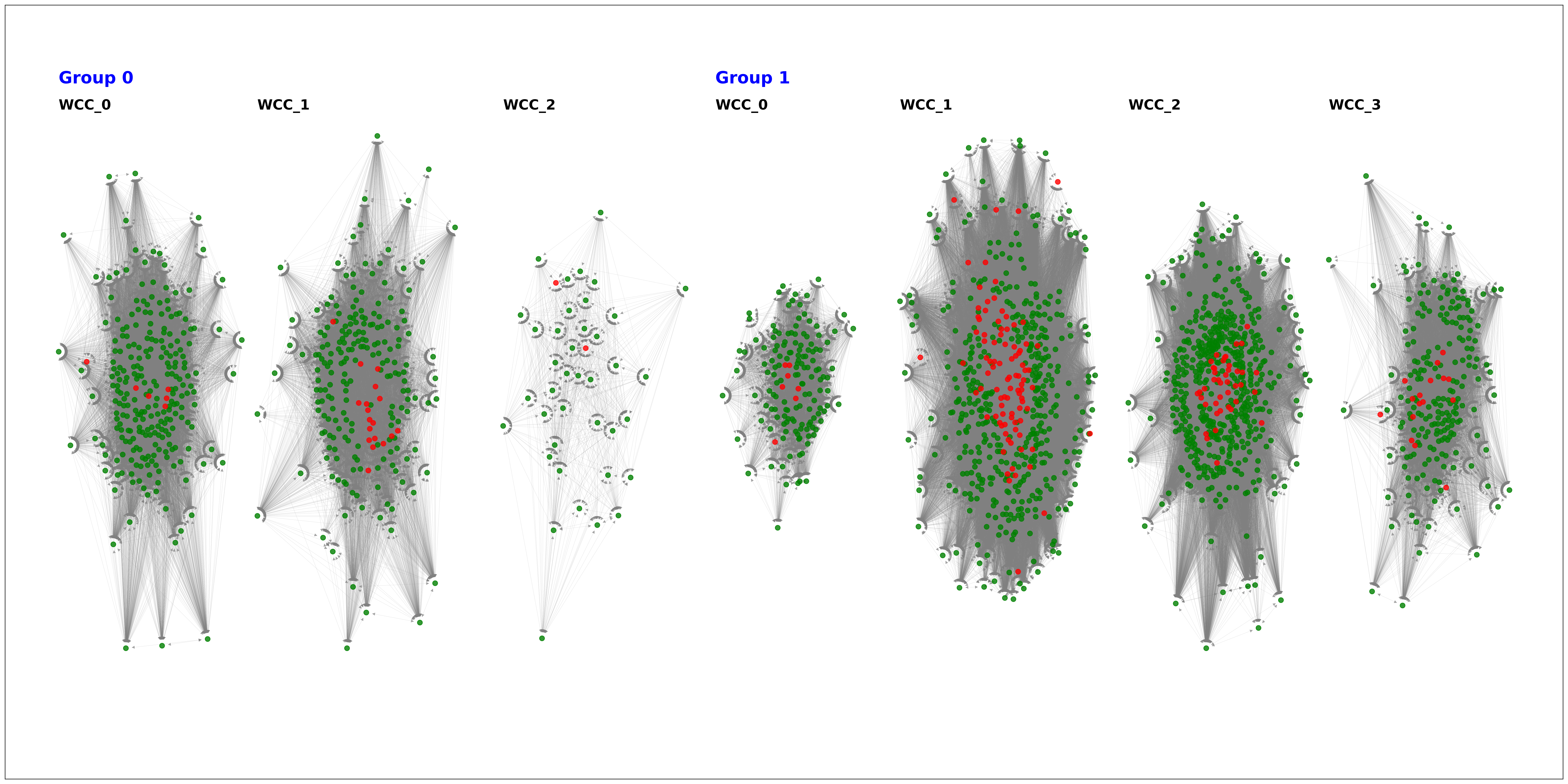}
    \caption{Visualization of the feasibility graph for the \texttt{COMPAS} dataset.}
    \label{fig: feasibility_graph}
\end{figure}

To illustrate the structure of the feasibility graph used in our approach, Figure~\ref{fig: feasibility_graph} presents a visualization constructed for the \texttt{COMPAS} dataset. In this graph, nodes correspond to data points and edges represent transitions that are feasible according to both real-world constraints (such as immutability or monotonicity of attributes) and plausibility constraints, which require that only small changes, those with cost below the $\epsilon$ threshold, are permitted.
This visualization highlights how these constraints shape the connectivity of the graph and naturally induce subgroup partitions.

\begin{longtable}{|p{3cm}|p{6cm}|p{0.6cm}|p{1.8cm}|}
    \caption{Attributes, descriptions, feasibility constraints, and data types for the \texttt{Adult}, \texttt{AdultCA}, \texttt{AdultLA}, \texttt{Student}, \texttt{COMPAS}, \texttt{German Credit}, and \texttt{HELOC} datasets. Feasibility Constraints (FC) are denoted as follows: Up arrows ($\uparrow$) indicate attributes, where only increases are allowed, down arrows ($\downarrow$), indicate attributes where only decreases are allowed, equal signs ($=$) denote attributes with no allowed changes, and dashes (-) represent attributes with no constraints.} \label{dataset: datasets} \\
    \hline
    \textbf{Attribute} & \textbf{Description} & \textbf{FC} & \textbf{Dtype} \\
    \hline \hline
    \endfirsthead
    
    \multicolumn{4}{c}{\textit{Continued from previous page}} \\
    \hline
    \textbf{Attribute} & \textbf{Description} & \textbf{FC} & \textbf{Dtype} \\
    \hline \hline
    \endhead
    
    \hline
    \multicolumn{4}{|r|}{\textit{Continued on next page}} \\
    \endfoot
    
    \hline
    \endlastfoot
    
    \multicolumn{4}{|c|}{\textbf{Adult Dataset}} \\
    \hline \hline
    age & Age of an individual & $\uparrow$ & \multirow{10}{*}{int64} \\
    workclass & Employment status of individual & - & \\
    education & Highest level of education attained & $\uparrow$ & \\
    educational-num & Number of years of education & $\uparrow$ & \\
    marital-status & Marital status & - & \\
    occupation & Occupation of individual & - & \\
    relationship & Relationship to the head of household & - & \\
    capital-gain & Capital gains in the past year & - & \\
    capital-loss & Capital losses in the past year & - & \\
    hours-per-week & Hours worked per week & - & \\
    \hline
    sex & Sex of individual & $=$ & \multirow{2}{*}{category} \\
    race & Race of individual & $=$ & \\

    \hline \hline
    \multicolumn{4}{|c|}{\textbf{AdultCA Dataset}} \\
    \hline \hline
    age & Age of an individual & $\uparrow$ & \multirow{4}{*}{int64} \\
    Class of worker & Employment status of individual & - & \\
    Educational Attainment & Highest level of education attained & $\uparrow$ & \\
    Marital Status & Marital status & - & \\
    Occupation & Occupation of individual & - & \\
    Place of Birth & Country or region where the individual was born & $=$ & \\
    Hours Worked per Week & Hours worked per week & - & \\
    \hline
    sex & Sex of individual & $=$ & \multirow{2}{*}{category} \\
    race & Race of individual & $=$ & \\

    \hline \hline
    \multicolumn{4}{|c|}{\textbf{AdultLA Dataset}} \\
    \hline \hline
    age & Age of an individual & $\uparrow$ & \multirow{10}{*}{int64} \\
    Class of worker & Employment status of individual & - & \\
    Educational Attainment & Highest level of education attained & $\uparrow$ & \\
    Marital Status & Marital status & - & \\
    Occupation & Occupation of individual & - & \\
    Place of Birth & Country or region where the individual was born & $=$ & \\
    Hours Worked per Week & Hours worked per week & - & \\
    \hline
    sex & Sex of individual & $=$ & \multirow{2}{*}{category} \\
    race & Race of individual & $=$ & \\

    \hline \hline
    \multicolumn{4}{|c|}{\textbf{Student Dataset}} \\
    \hline \hline
    age &  Age of student (15 to 22)& $\uparrow$ & \multirow{17}{*}{int64}\\
    Medu & Education of mother & $\uparrow$ & \\
    Fedu & Education of father & $\uparrow$ & \\
    traveltime & Home to school travel time & - & \\
    studytime & Weekly study time & - & \\ 
    failures & Number of past class failures & - & \\
    famrel & Quality of family relationships & - & \\
    freetime & Free time after school & - & \\
    goout & Going out with friends & - & \\
    Dalc & Workday alcohol consumption & - & \\
    Walc & Weekend alcohol consumption & - & \\
    health & Current health status & $\uparrow$ & \\
    absences & Number of school absences & - & \\
    G1 & First period grade & - & \\
    G2 & Second period grade & - & \\
    G3 & Final grade & - & \\
    target & $1 \text{ if } G3 \geq 10 \text{ else } 0$ & - & \\
    \hline
    school & School of student & - & \multirow{3}{*}{category} \\ 
    sex & Sex of student & $=$ &  \\
    address & Home address type of student & - & \\
    famsize & Family size & $\uparrow$ & \\
    Pstatus & Cohabitation status of parents& - & \\
    Mjob & Job of mother & - & \\
    Fjob & Job of father & - & \\
    reason & Reason to choose this school& - & \\
    guardian & Guardian of student & - & \\
    schoolsup & Extra educational support & - & \\
    famsup & Family educational support& - & \\
    paid & Extra paid classes within the course subject& - & \\
    activities & Extra-curricular activities & - & \\
    nursery & Attended nursery school & $\uparrow$ & \\
    higher & Wants to take higher education & - & \\
    internet & Internet access at home & - & \\
    romantic & with a romantic relationship & - & \\

    \hline \hline
    \multicolumn{4}{|c|}{\textbf{Compas Dataset}} \\
    \hline \hline
    age & Age of defendant & $\uparrow$ & \multirow{5}{*}{int64} \\
    juv\_fel\_count & Juvenile felony count  &$\downarrow$ &  \\
    juv\_misd\_count & Juvenile misdemeanor count  &$\downarrow$ &  \\
    juv\_other\_count & Juvenile other offenses count  &$\downarrow$ &  \\
    priors\_count & Prior offenses count  &$\downarrow$ &  \\  
    \hline
    sex & Sex of defendant & = & \multirow{5}{*}{category} \\
    c\_charge\_degree &  Charge degree of original crime  & $\downarrow$ & \\
    race & Race of defendant & - &  \\
    two\_year\_recid & Whether the defendant is rearrested within 2 years & - &  \\

    \hline \hline
    \multicolumn{4}{|c|}{\textbf{German Credit Dataset}} \\
    \hline \hline
    Credit-Amount  & Amount of credit required& $\downarrow$ & Continuous \\
    \hline
     Month-Duration & Duration of the credit in months & $\downarrow$ & \multirow{8}{*}{int64} \\
     Installment-Rate & Installment rate as a percentage of disposable income& $\downarrow$ & \\
     Residence & Present residence duration in years& - & \\
     Age  & Age of the individual& $\uparrow$ & \\
     Existing-Credits & Number of existing credits at this bank& - & \\
     Num-People & Number of people liable to provide maintenance& - & \\
    \hline
     Existing-Account-Status&Balance or type of the checking account& $\uparrow$ & \multirow{6}{*}{category} \\
     Credit-History& Past credit behavior of individual & - & \\
     Purpose& Purpose of the credit (e.g., furniture, education) & - & \\
     Savings-Account&Status of savings account/bonds  & $\uparrow$ & \\
     Present-Employment& Duration of present employment & $\uparrow$ & \\
     Sex& Sex of the applicant & $=$ & \\
     Marital-Status& Marital status of the applicant & $=$ & \\
     Guarantors& Presence of guarantors & $\uparrow$ & \\
     Property& Property ownership & $\downarrow$ & \\
     Installment&Other installment plans  & $\downarrow$ & \\
     Housing&Housing status (e.g., rent, own, for free)  & $\uparrow$ & \\
     Job& Job type (e.g., unemployed, management) & $\uparrow$ & \\
     Telephone& Registered telephone under the customers name & $\uparrow$ & \\
     Foreign-Worker&Whether the applicant is a foreign worker & $=$ & \\

    \hline \hline
    \multicolumn{4}{|c|}{\textbf{HELOC Dataset}} \\
    \hline \hline
     AverageMInFile & Average months in file for all trade lines & - & \multirow{20}{*}{Continuous} \\
     NetFraction Install Burden & Net fraction of installment credit to credit limit &$\downarrow$ &  \\
     NetFraction Revolving Burden & Net fraction of revolving credit to credit limit & - &  \\
    MSinceMostRecent Trade Open & Months since the most recent trade line was opened & - &  \\
    PercentInstall Trades & Percentage of installment trades & - &  \\
     PercentTrades WBalance & Percentage of trades with balance & - &  \\ 
     NumTotalTrades & Total number of trade lines & -  &  \\
     MSinceMostRecent Delq & Months since the most recent delinquency &$\downarrow$ &  \\
     NumSatisfactory Trades & Number of satisfactory trade lines &$\uparrow$ &  \\
     PercentTradesNever Delq & Percentage of trades with no delinquency &$\uparrow$ &  \\ 
     ExternalRisk Estimate & Risk estimate provided by an external source &$\downarrow$ & \\
     \hline
     ExternalRisk Estimate & Risk estimate provided by an external source &$\downarrow$ & \multirow{6}{*}{int64} \\
     MSinceOldest TradeOpen & Months since the oldest trade was opened & - &  \\
    NumTrades60Ever2 DerogPubRec & Number of trades that have experienced 60+ days past due or worse &$\downarrow$ &  \\
     NumTrades 90Ever 2DerogPubRec & Number of trades that have experienced 90+ days past due or worse &$\downarrow$ &  \\
     MaxDelq2Public RecLast12M & Maximum delinquency reported in the last 12 months &$\downarrow$ &  \\
     MaxDelqEver & Maximum delinquency reported ever &$\downarrow$ &  \\
     NumTradesOpenin Last12M & Number of trades opened in the last 12 months & - &  \\
     MSinceMostRecent Inqexcl7days & Months since the most recent inquiry, excluding the last 7 days & - &  \\
     NumInqLast6M & Number of inquiries in the last 6 months & - &  \\
    NumInqLast6 Mexcl7days & Number of inquiries in the last 6 months, excluding the last 7 days & - &  \\
     NumRevolving Trades WBalance & Number of revolving trades with balance & - &  \\
    NumInstallTrades WBalance & Number of installment trades with balance & - &  \\
     NumBank2Natl Trades WHigh Utilization & Number of bank/national trades with a high utilization ratio & - &  \\ \hline
     RiskPerformance & Target variable indicating borrower's risk performance & - & \multirow{2}{*}{cateegory} \\
\end{longtable}

\subsection{Feasibility Graph}
\begin{figure}[ht]
    \centering
    \begin{subfigure}[t]{0.42\textwidth}
        \centering
        \includegraphics[width=\linewidth]{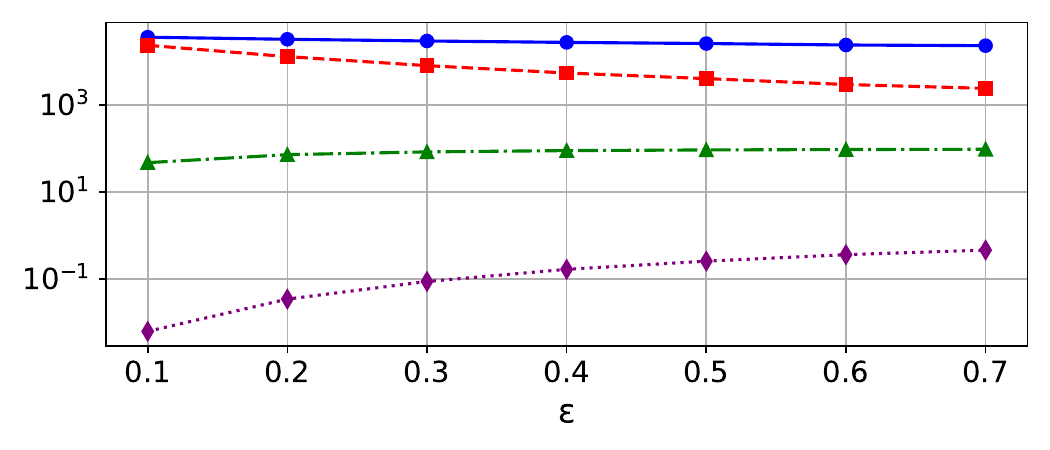}
        \caption{\texttt{AdultCA} Dataset}
    \end{subfigure}
    \hfill
    \begin{subfigure}[t]{0.42\textwidth}
        \centering
       \includegraphics[width=\linewidth]{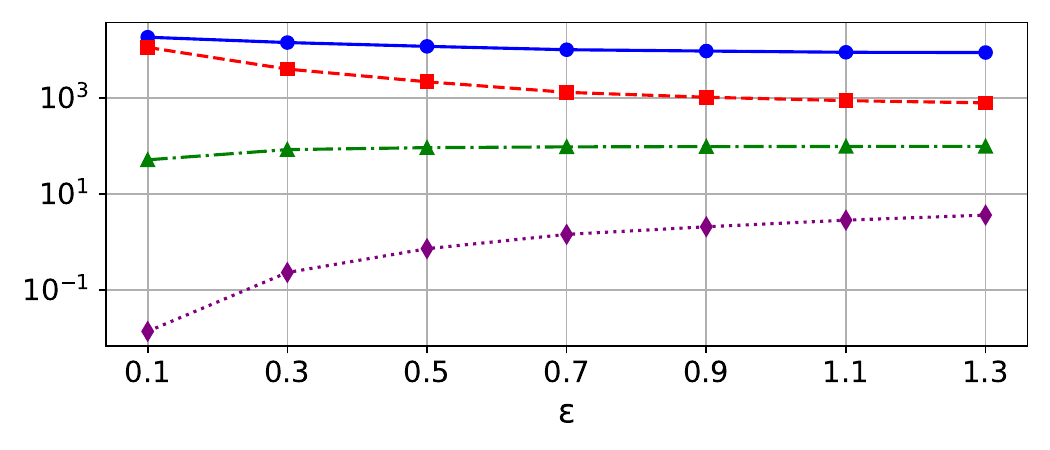}
       \caption{\texttt{AdultLA} Dataset}
    \end{subfigure}
    \hfill
    \begin{subfigure}[t]{0.42\textwidth}
        \centering
       \includegraphics[width=\linewidth]{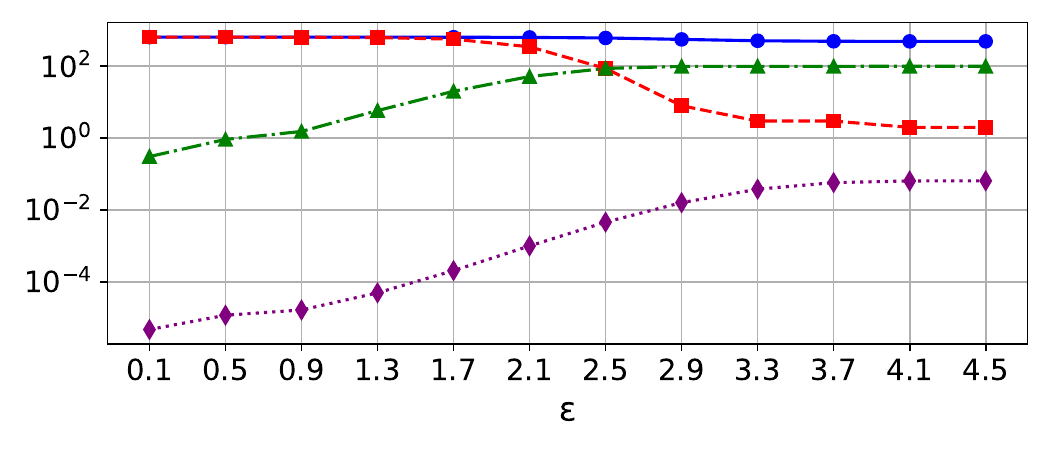}
        \subcaption{\texttt{Student} Dataset}
    \end{subfigure}
    \hfill
    \begin{subfigure}[t]{0.42\textwidth}
        \centering
       \includegraphics[width=\linewidth]{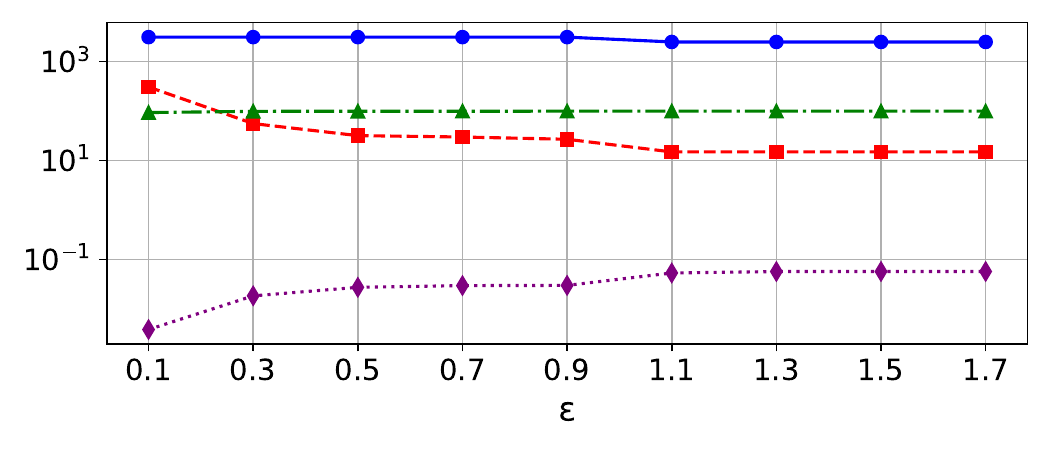}
        \subcaption{\texttt{COMPAS} Dataset}
    \end{subfigure}
    \hfill
    \begin{subfigure}[t]{0.42\textwidth}
        \centering
       \includegraphics[width=\linewidth]{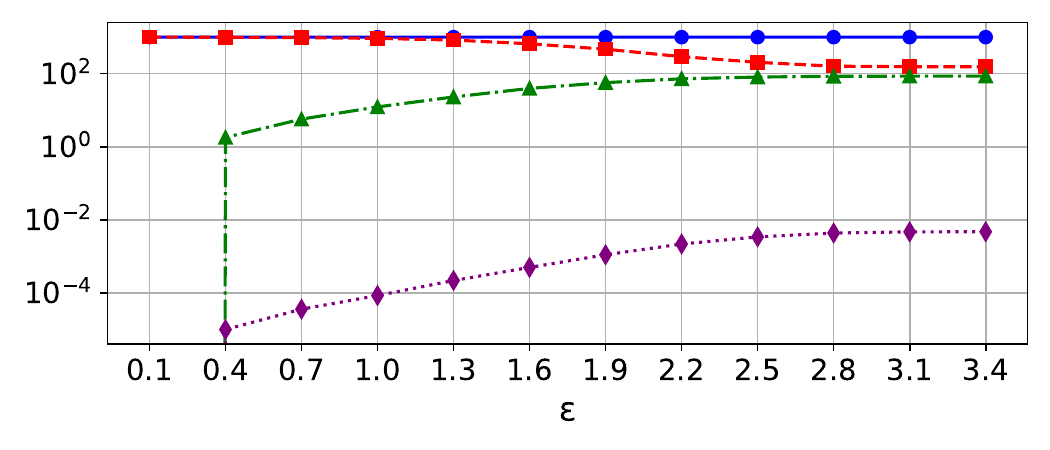} 
        \subcaption{\texttt{German Credit} Dataset}
    \end{subfigure}
    \hfill
    \begin{subfigure}[t]{0.42\textwidth}
        \centering
       \includegraphics[width=\linewidth]{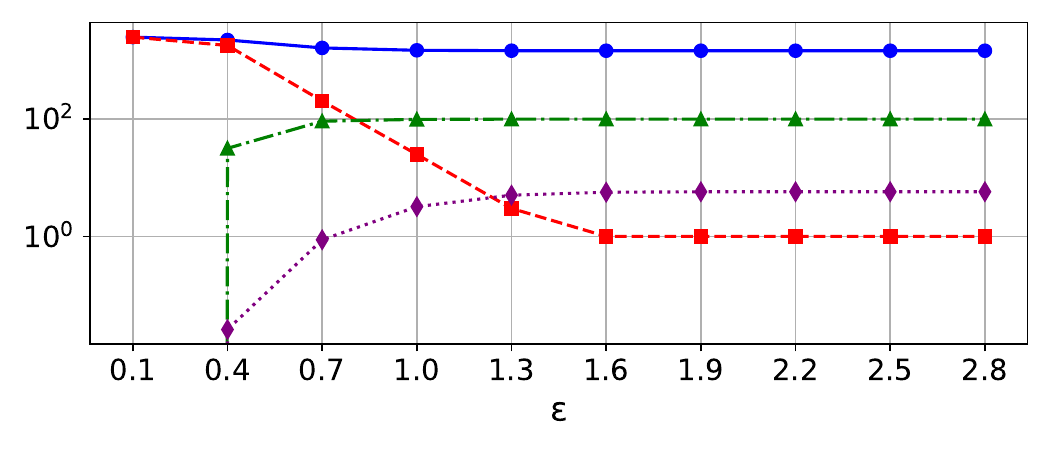} 
        \subcaption{\texttt{HELOC} Dataset}
    \end{subfigure}
    \hfill
    \begin{subfigure}{0.35\textwidth}
        \raisebox{0.4cm}{
    \includegraphics[width=\linewidth]{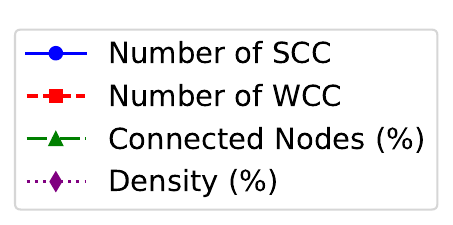}}
    \end{subfigure}
\captionsetup{justification=centering}
    \caption{Feasibility graph connectivity based on $\epsilon$.}
    \label{fig: decide_e_appendix}
\end{figure}

We determine dataset-specific $\epsilon$ values by balancing plausibility and connectivity: $0.7$ for \texttt{AdultCA}, $0.5$ for \texttt{AdultLA}, $3$ for \texttt{Student}, $1.1$ for \texttt{COMPAS}, $3.1$ for \texttt{German Credit}, and $1.3$ for \texttt{HELOC} (Figure~\ref{fig: decide_e_appendix}).
For all the datasets except the \texttt{HELOC}, we define groups based on gender, with $G_0$ representing women and $G_1$ representing men. In \texttt{HELOC}, we use the \textit{MaxDelqEver} attribute to distinguish between individuals with more than five delinquencies ($G_1$) and those with five or fewer ($G_0$).

\section{Implementation Details}
\label{seq: implementation_details}
In this section, we present the implementation details of our algorithms and metrics. We describe the models used, the dataset preprocessing pipeline, the density selection method for constructing the feasibility graph, and the key parameters employed in our methods.

\textbf{Models.} Our experiments employ the Logistic Regression model\footnote{\href{https://scikit-learn.org/stable/modules/generated/sklearn.linear_model.LogisticRegression.html}{Logistic Regression from scikit-learn}} from scikit-learn\footnote{\href{https://scikit-learn.org/stable/index.html}{Python package scikit-learn}}, utilizing its default settings for classification tasks. The datasets are divided into training and test sets with a 70\% to 30\% ratio, respectively. For reproducibility, we use a random seed value of 482 in the `train\_test\_split` function\footnote{\href{https://scikit-learn.org/stable/modules/generated/sklearn.model_selection.train_test_split.html}{train\_test\_split python package scikit-learn}} from scikit-learn. FACEGroup is applied exclusively to the test set.

\textbf{Preprocess.}
Our preprocessing pipeline is consistent across datasets, including the treatment of categorical and numerical data, which often poses challenges in these tasks.
For ordered categorical attributes, we allow for a user-defined order that is dataset-specific, offering flexibility to accommodate natural sequences where applicable. In contrast, unordered categorical attributes, such as job occupations, are one-hot encoded to ensure that the difference between any two categories is consistent and maximally distinct without implying any order or magnitude of change. Binary categorical attributes are label-encoded.
For continuous attributes, we discretize them into bins using a heuristic algorithm that determines the number of bins based on the logarithm of the number of unique values and the range of those values, ensuring a balance between granularity and practicality. 
Last, we normalize each attribute to the range of $[0,1]$, ensuring an equal contribution of attributes. 

Although some datasets require additional tailored adjustments.
From the \texttt{COMPAS} dataset, we exclude the attributes \textit{age\_cat} and \textit{c\_charge\_desc}. The \textit{age\_cat} attribute, which bins ages into categories, was excluded due to its strong correlation coefficient of $0.99$ with \textit{age}. Similarly, the \textit{c\_charge\_desc} attribute was omitted because of its high correlation coefficient of $0.91$ with \textit{c\_charge\_degree}, a binary attribute indicating misdemeanor or felony status. Retaining \textit{age\_cat} and \textit{c\_charge\_degree} simplifies handling binary attributes.
For the \texttt{German Credit} dataset, we split the original sex attribute, which contains information about sex and marital status, into two distinct attributes: sex and marital status \cite{le2022survey}. 
To create feasibility constraints, we adjust certain attribute values as follows: For the \textit{Existing-Account-Status} attribute, we map the value 'A14' to 'A10 ' so that account status can only improve in the proposed counterfactual instance after encoding.
Similarly, for the \textit{Savings account/bonds} attribute, we map 'A65' to 'A60' to impose constraints that allow only increases in savings in the proposed counterfactual instance. 
Additionally, for the \textit{Credit-History} attribute, we set the constraint that if the instance value is A34 or A33, it can change to A32, A31, or A30, ensuring an improved credit history while permitting all other possible transitions.
For the \texttt{HELOC} dataset, we drop any row with at least one negative value in its columns.

\textbf{Density Selection Method.} 
Bandwidth selection is a critical aspect of kernel density estimation (KDE) that 
impacts the smoothness and accuracy of the resulting density estimate. To address this, we offer two methods for bandwidth selection. The first method is based on the rule-of-thumb \cite{scott2015multivariate}, a heuristic that balances bias and variance in density estimation. It calculates the bandwidth parameter using the sample size and standard deviation of the data, providing a simple yet effective way to determine bandwidth. Alternatively, for a more refined optimization, we calculate the bandwidth using grid search and cross-validation techniques.

\textbf{MIP Algorithm.}
We implement the MIP approach using the Pulp library \cite{mitchell2011pulp}, with the COIN-OR Branch and Cut (CBC) solver, which executes the branch-and-cut algorithm.

\textbf{Maximum Possible Cost.}\label{max_possible_cost}
In our experiments, the maximum possible cost refers to the maximum cost between all pairs in the dataset.

\textbf{AUC Scores Parameters Settings.}
For \( kAUC \), \( dAUC \), and \( cAUC \), we evaluate each metric using four evenly spaced input values.
Specifically, \( kAUC \) is computed over four values in the range \([1, k_0]\), \( dAUC \) uses four cost values from \([0.1, \text{maximum possible cost}]\), and \( cAUC \) is measured at four percentage coverage levels up to full coverage (\( c = 1 \)).
The integral computation of these metrics spans different ranges on the x-axis. Without loss of generality, for \( kAUC \), integration is performed over 12 steps of the cost range \([0.1, \text{maximum possible cost}]\). For \( dAUC \), the integral is computed across $12$ counterfactual values in \([1, k_0]\). Finally, for \( cAUC \), the integral spans cost values across $25$ values of \([1, k_0]\).
For the \texttt{Adult} dataset in the main paper, we set the maximum possible cost to \( 2 \), considering it sufficiently large.

To ensure evenly spaced and interpretable divisions for our $AUC$ metrics and plots, we adopt a \textit{nice numbers} approach inspired by \cite{Talbot2010AnEO}. This method refines raw tick spacing by selecting values that align with intuitive numerical scales, improving readability.
Given a range $[\text{min}, \text{max}]$ and a desired number of intervals $n$, the initial spacing is computed as:
\(
\text{tick\_spacing} = \frac{\text{max} - \text{min}}{n - 1}.
\label{alg: nice_numbers}
\)
which is adjusted based on its order of magnitude and rounded to the nearest \textit{nice} number. For integers like the number of counterfactuals, we select from $\{1, 2, 3, 4, 5, 7, 10\}$, while for decimals, we select from the whole range of $1$ to $10$ and scale it accordingly.

\textbf{Optimal $kAUC$ and $dAUC$.}
The optimal $kAUC$ and $dAUC$ scores are computed as the AUC of the values along the x-axis (number of counterfactuals $k$ for $kAUC$, or costs $d$ for $d$AUC) while maintaining the coverage at its maximum. 
These optimal scores provide a baseline for normalization, ensuring that the $kAUC$ and $dAUC$ scores reflect the best achievable performance across all $k$ or $d$ values, respectively.

\textbf{Face Comparison.}
To ensure a fair comparison between FACEGroup and FACE, we normalize both vector costs and shortest path costs by the maximum observed value across all instances.
Also, the number of counterfactuals as input is determined by dividing the range $[1, k_0]$ into ten evenly spaced values based on \ref{alg: nice_numbers}.

\textbf{Baselines Comparison.}
To compare our approach with baselines, for FACE and GLOBE-CE, we utilize their code repositories available on GitHub. For AReS, we employ the code provided by GLOBE-CE as no dedicated repository exists.

\section{Additional Experiments}
\label{seq: additional_experiments}
In this section, we provide the minimum counterfactual resources ($k_0, d_0$) required for full coverage for the additional datasets and further compare our approach with FACE.
Additionally, we provide a performance comparison between our Greedy and MIP solutions for both problems.
\subsection{Minimum Resources}
Table \ref{tab: d0k0_appendix} reports the counterfactual burden in terms of the minimum resources needed for full coverage per group.
 Unlike the subgroup-level analysis in the main paper, these results aggregate the overall counterfactual burden for each group.

Across datasets, $G_1$ generally requires more counterfactuals ($k_0$), suggesting that feasible transitions for this group are more dispersed. However, \texttt{HELOC} exhibits the opposite trend.
In contrast, $G_0$ tends to have higher minimum cost thresholds ($d_0$) across datasets, except for \texttt{AdultLA} and \texttt{COMPAS}, where $G_1$ exhibits greater cost requirements. These variations highlight dataset-specific differences in the structure of feasible transitions, which influence the counterfactual burden across groups.
\begin{table}[H]
    \centering
    \setlength{\tabcolsep}{8pt}
    \captionsetup{justification=centering}
    \caption{Minimum counterfactuals and cost for coverage.}
    \begin{tabular}{lcc|cc}
    \hline
        \multirow{2}{*}{Datasets} & \multicolumn{2}{c}{$G_0$} & \multicolumn{2}{c}{$G_1$} \\ \cmidrule{2-5}
        & $k_0$ & $d_0$ & $k_0$ & $d_0$ \\
        \hline
        \texttt{AdultCA} & 58 & 1.85 & 75 & 1.05\\
        \texttt{AdultLA} & 24 & 1.15 & 35 & 1.28\\
        \texttt{Student}  & 2 & 3.61 & 2 & 3.59 \\ 
        \texttt{COMPAS} & 3 & 1.09 & 4 & 1.22\\
        \texttt{German Credit} & 4 & 2.50 & 6 & 2.43\\
        \texttt{HELOC} & 6 & 1.55 & 1 & 1.54\\
        \hline
    \end{tabular}
    \captionsetup{justification=centering}
    \label{tab: d0k0_appendix}
\end{table}

\subsection{Comparison with Individual Counterfactuals}
\begin{figure}[h]
    \centering
    \begin{subfigure}{.48\textwidth}
        \centering
        \includegraphics[width=\linewidth]{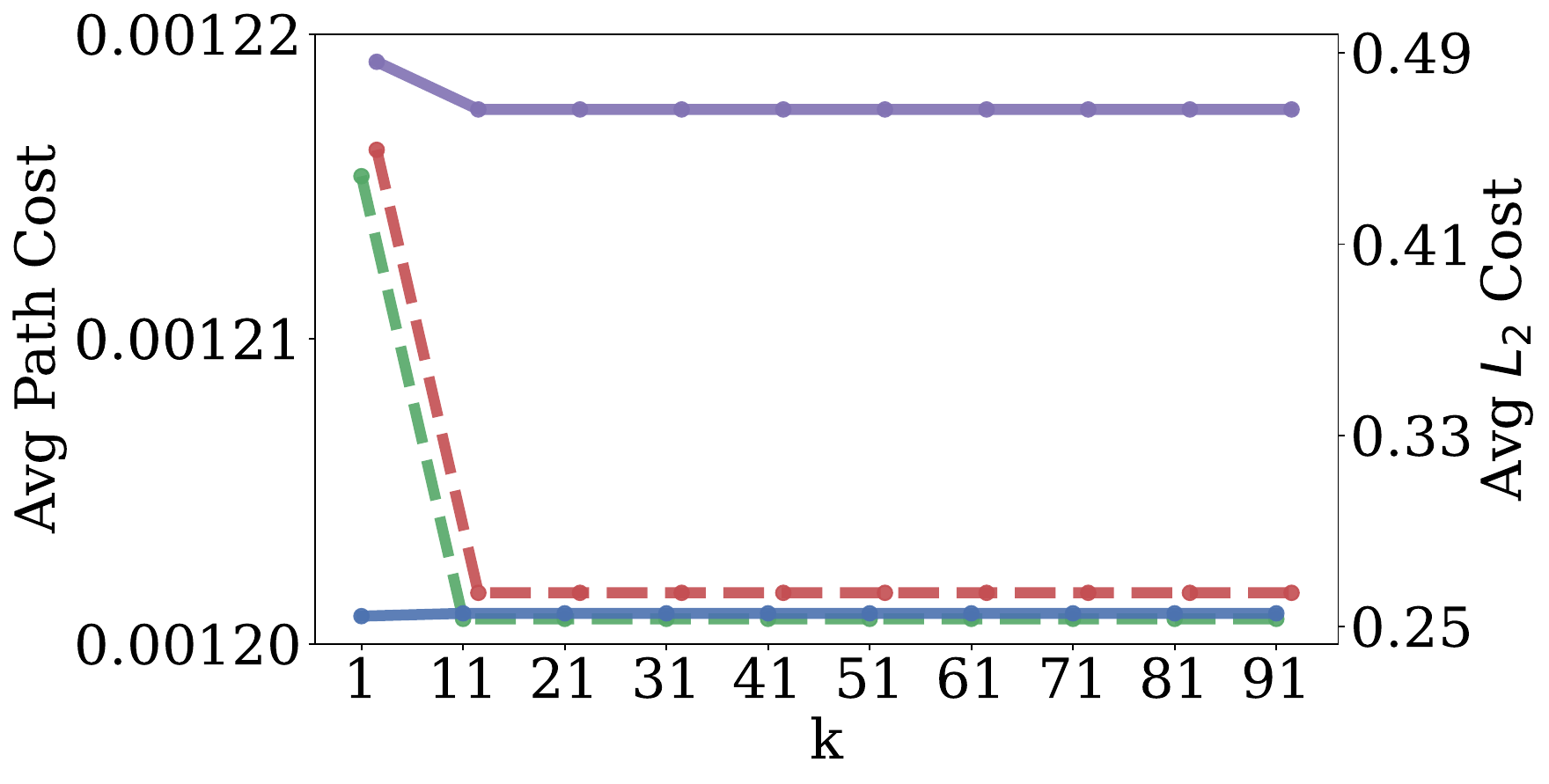}
        \subcaption{\texttt{AdultCA} Dataset}
    \end{subfigure}
    \hfill
    \begin{subfigure}{.48\textwidth}
        \centering
        \includegraphics[width=\linewidth]{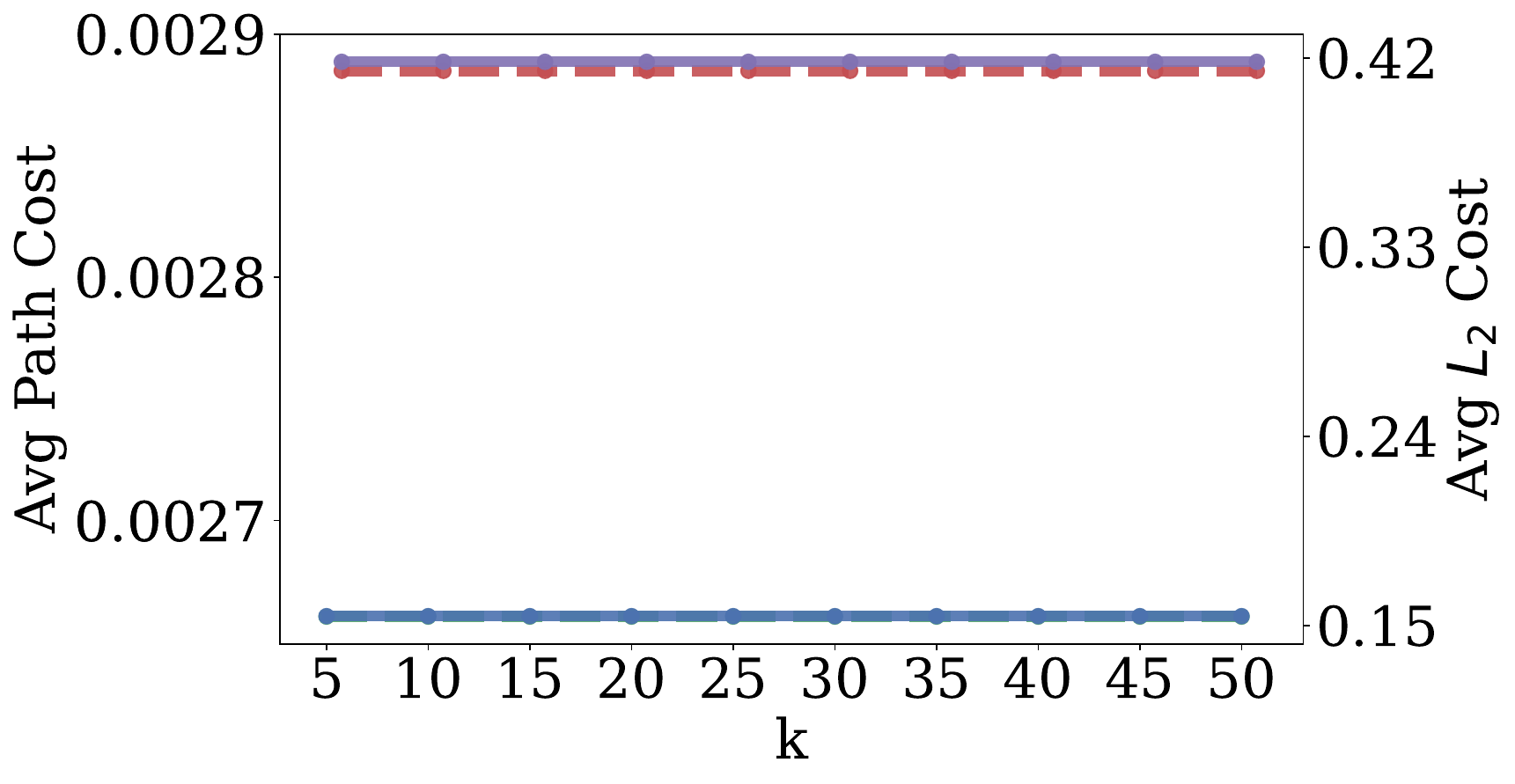}
        \subcaption{\texttt{AdultLA} Dataset}
    \end{subfigure}
    \vspace{0.5em}
    \begin{subfigure}{.48\textwidth}
        \centering
        \includegraphics[width=\linewidth]{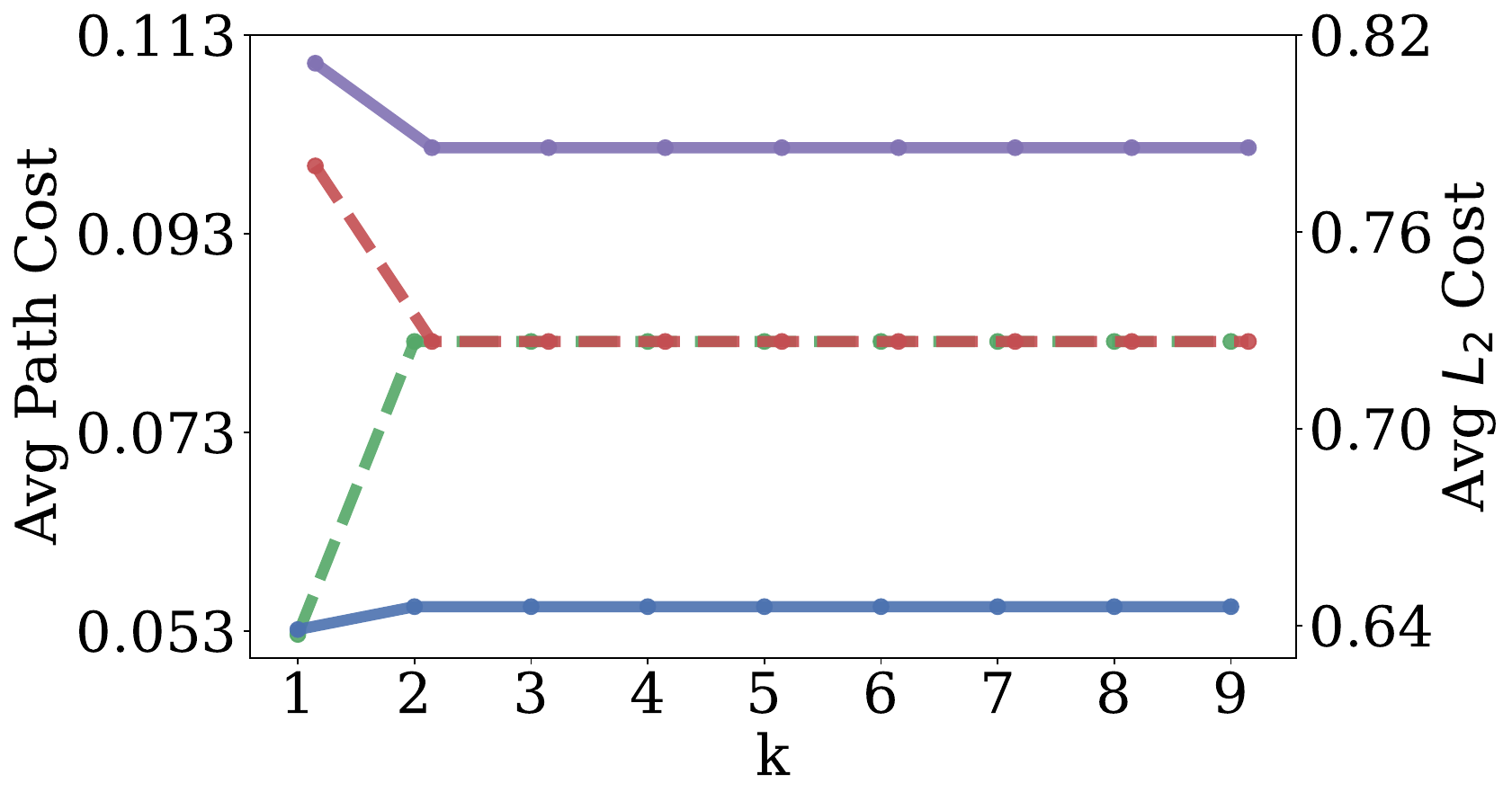}
        \subcaption{\texttt{Student} Dataset}
    \end{subfigure}
    \hfill
    \begin{subfigure}{.48\textwidth}
        \centering
        \includegraphics[width=\linewidth]{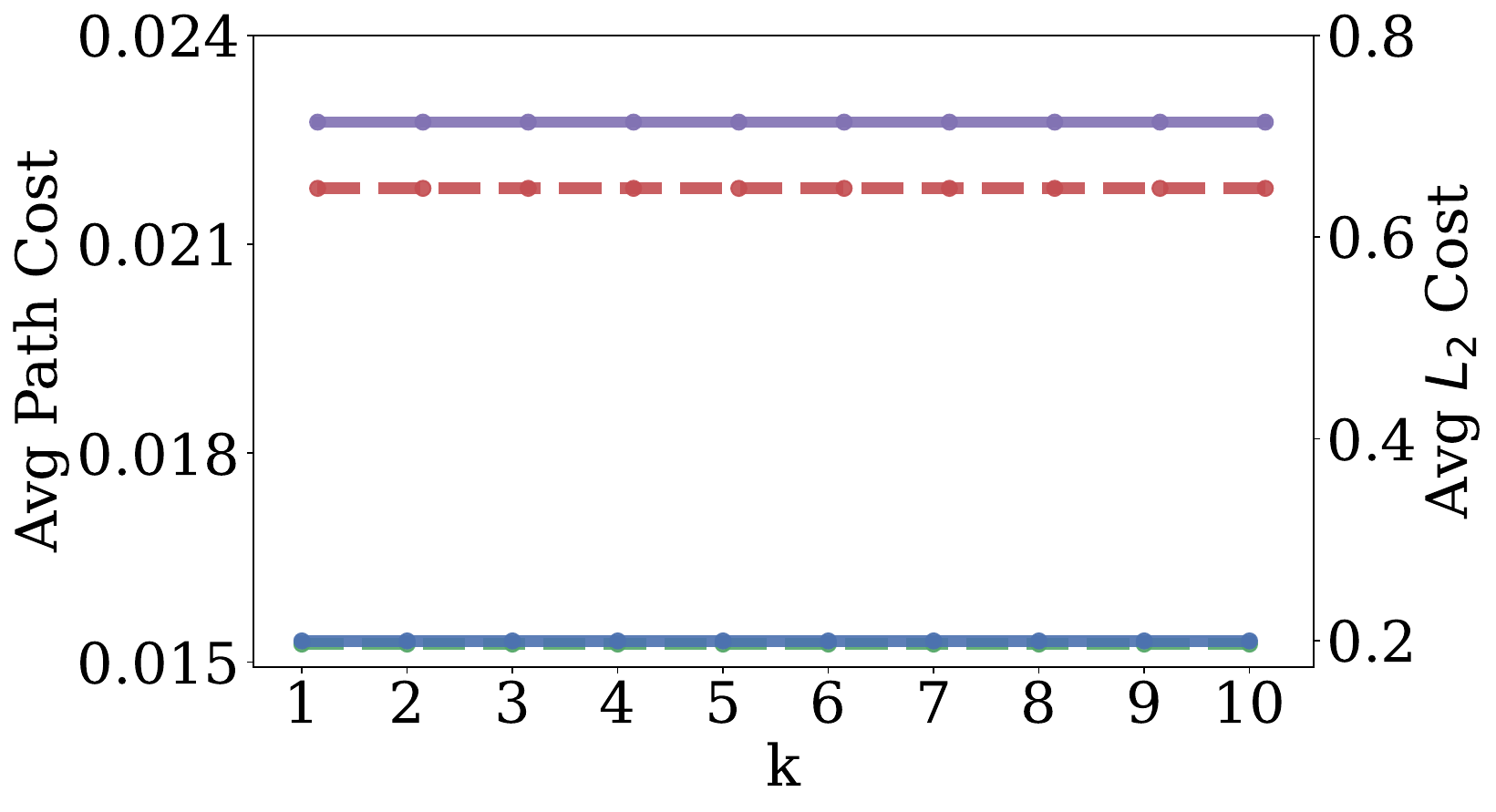}
        \subcaption{\texttt{COMPAS} Dataset}
    \end{subfigure}

    \vspace{1em}
    \includegraphics[width=0.95\textwidth]{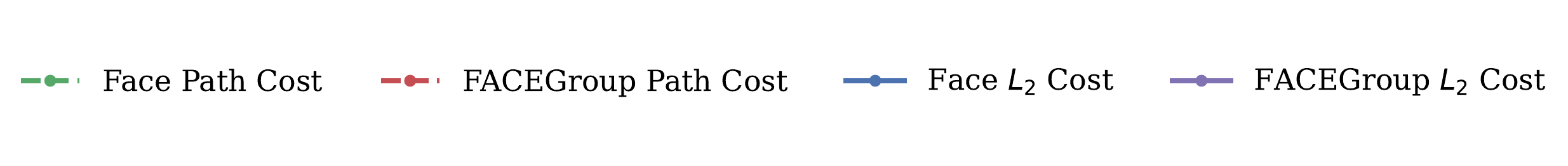}

    \captionsetup{justification=centering}
    \caption{Comparison of FACEGroup and FACE based on average CF costs.}
    \label{fig: face_appendix}
\end{figure}

To further assess the cost of FACEGroup to the optimal cost of individual CFs from FACE, we extend the evaluation to additional datasets (Figure~\ref{fig: face_appendix}). As in the main paper, we compare the average factual-to-counterfactual cost under two metrics: (a) weighted shortest path cost in $G_U$ and (b) pairwise $L_2$ cost. 

FACE consistently achieves the lowest costs, while FACEGroup incurs slightly higher costs due to its group-level constraints.
However, FACEGroup remains close to optimal in datasets like \texttt{AdultCA}, \texttt{AdultLA}, and \texttt{Student}, where feasible transformations align well with individual counterfactual selections.
In contrast, in \texttt{COMPAS}, the cost gap between FACE and FACEGroup widens, suggesting that group-level counterfactual selection requires higher-cost transitions to maintain coverage, as fewer low-cost feasible pathways exist within the feasibility graph.

\subsection{Coverage of Baseline Approaches without Constraints}
Table~\ref{tab:baselines_coverage} reports the coverage and number of GCFs generated without considering any feasibility constraints for baselines. AReS produces a compact set of interpretable rules, while GLOBE-CE generates a larger set due to its scalable translation framework. Both methods exceed 70 \% coverage across datasets.
\begin{table}[H]
\centering
\renewcommand{\arraystretch}{1.2}
\setlength{\tabcolsep}{6pt}
\caption{Baselines Coverages.}
\label{tab:baselines_coverage}
\begin{tabular}{lcccccc}
\toprule
Dataset & \multicolumn{2}{c}{AReS} & \multicolumn{2}{c}{GLOBE-CE} \\ 
\cmidrule(lr){2-3} \cmidrule(lr){4-5}
& r & Coverage (\%) & k & Coverage (\%)\\ 
\midrule
Adult & 18 & 96.43 & 421 & 99.76 \\
AdultCA & 20 & 81.06 & 612 & 85.83 \\
AdultLA & 20 & 86.61 & 342 & 83.21  \\
Student & 3 & 83.33 & 10 & 83.33  \\
COMPAS & 20 & 87.4 & 124 & 91.85  \\
German Credit & 4 & 78.94 & 18 & 94.73 \\
HELOC & 11 & 70 & 74 & 72.27 \\
\bottomrule
\end{tabular}
\end{table}

\subsection{Performance Comparison of Greedy and MIP Algorithms}
This section presents supplementary evaluations that support the main results. We provide further insights into feasibility graph construction, minimum counterfactual resources, and individual counterfactual comparisons. Additionally, we report baseline coverages without feasibility constraints.

We compare the performance of the greedy and Mixed Integer Programming (MIP) approaches for both cost and coverage-constrained counterfactual selection, evaluating their ability to balance computational efficiency and cost-effectiveness.

\begin{figure}[h!]
    \centering
    \begin{subfigure}{0.41\textwidth}
        \centering
        \includegraphics[width=\linewidth]{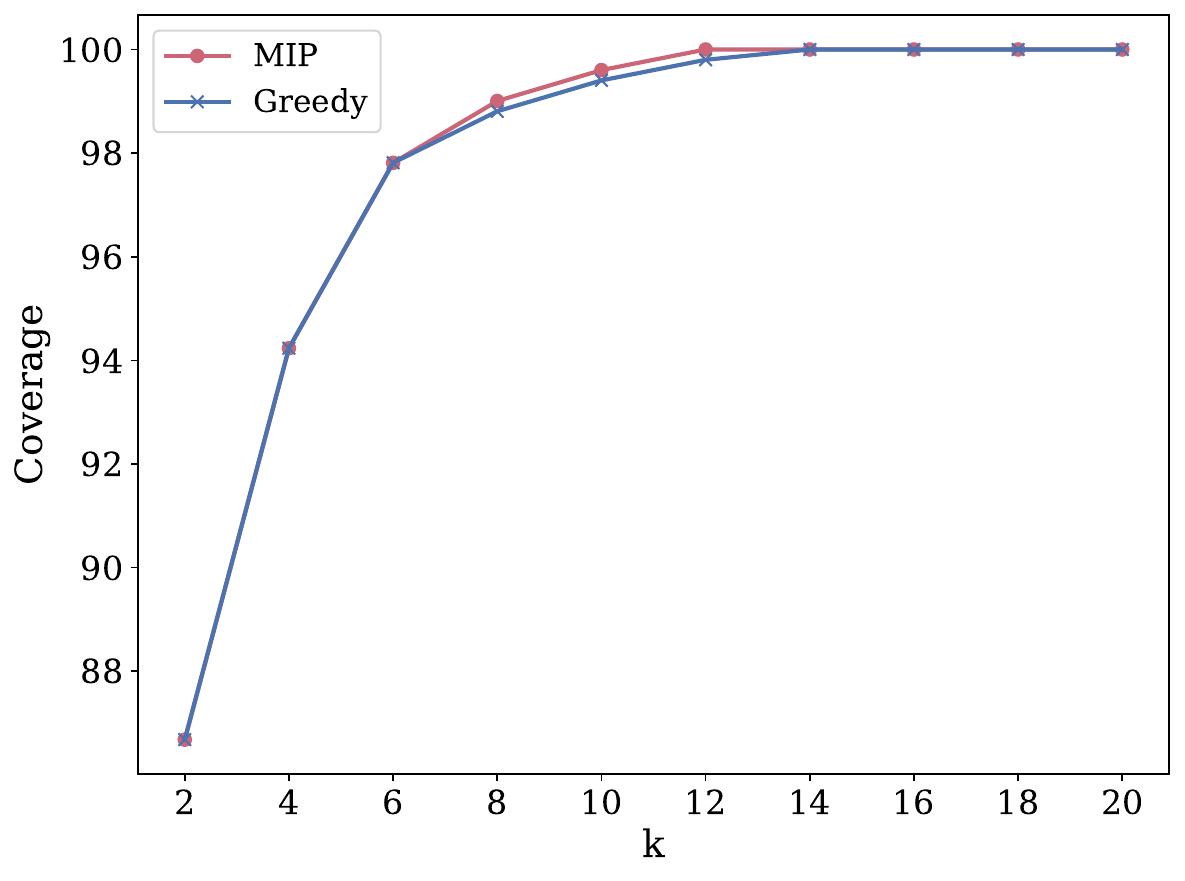}
        \caption{Total Coverage}
    \end{subfigure}
    \hfill
    \begin{subfigure}{0.41\textwidth}
        \centering
        \includegraphics[width=\linewidth]{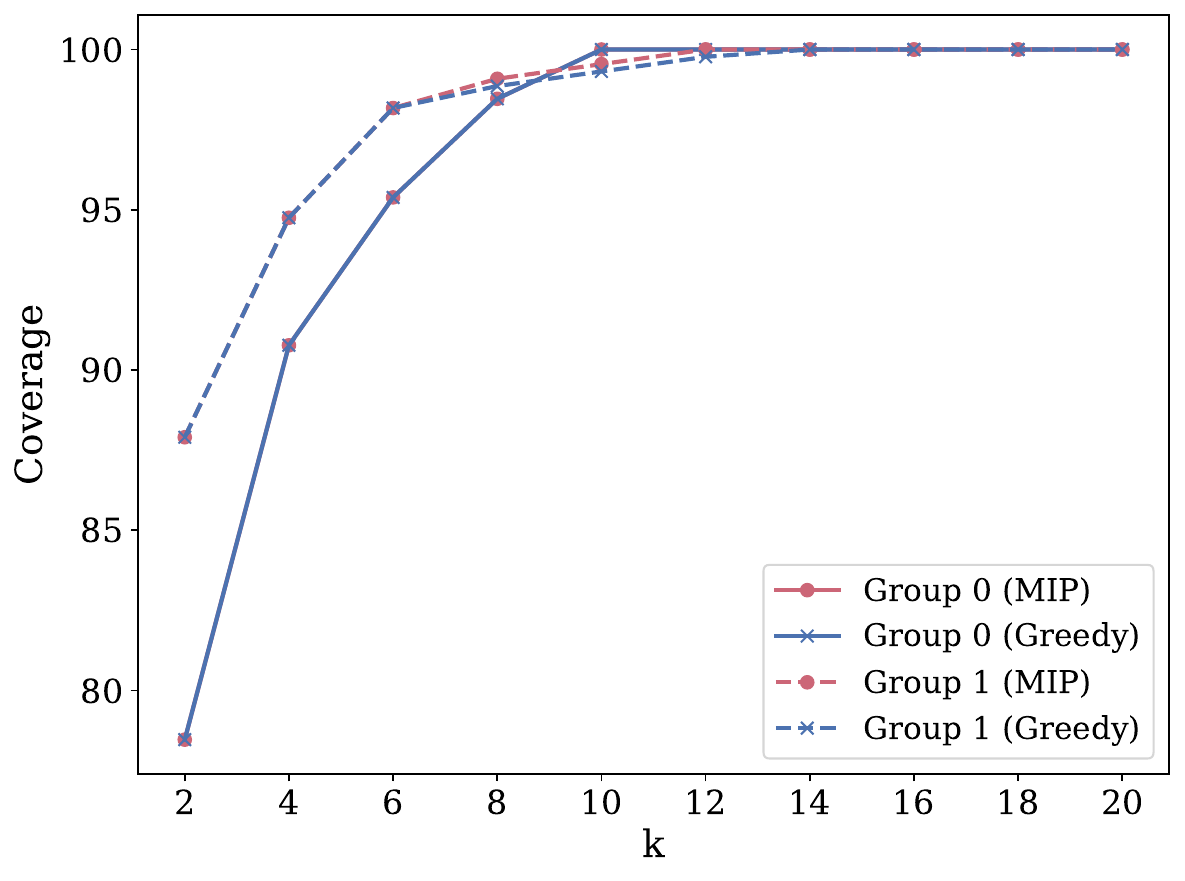}
        \caption{Group Coverage}
    \end{subfigure}
    \hfill
    \begin{subfigure}{0.41\textwidth}
        \centering
        \includegraphics[width=\linewidth]{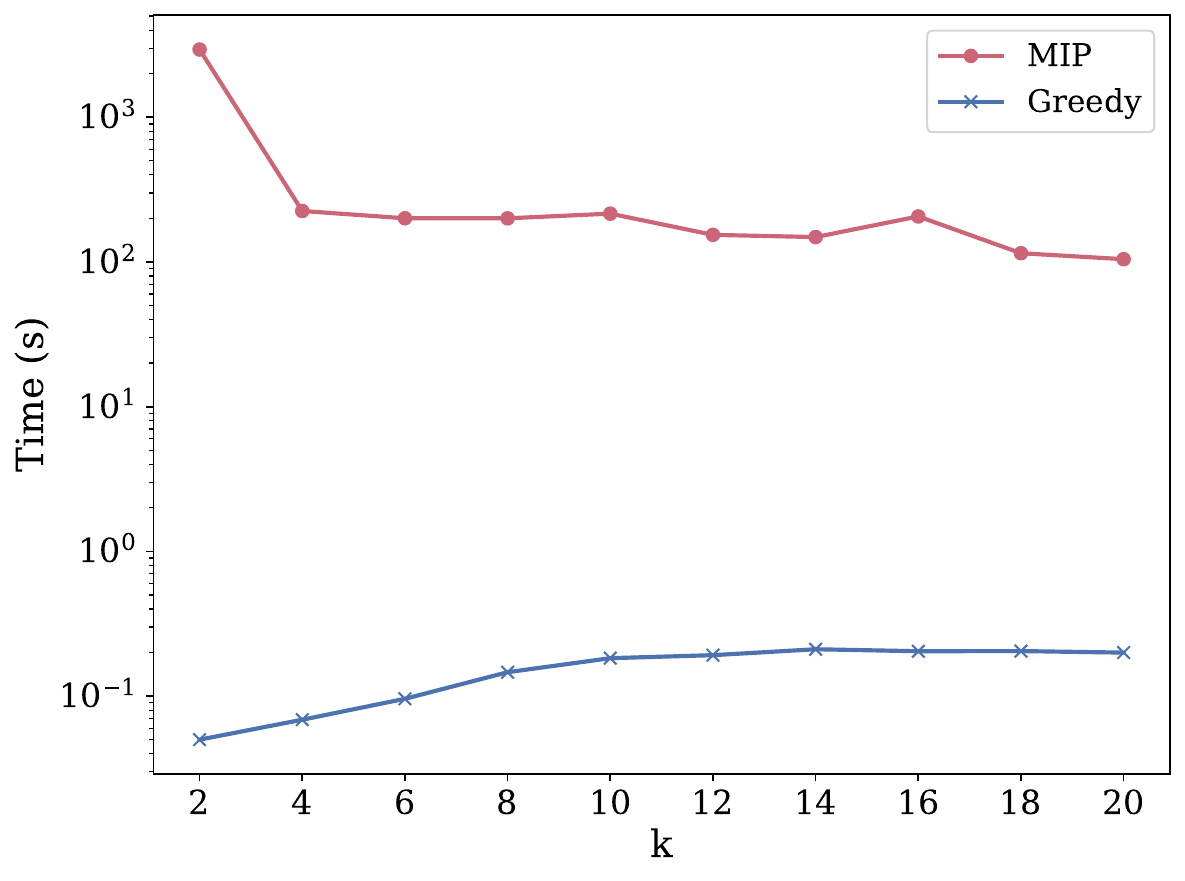} 
        \caption{Time}
    \end{subfigure}
    \hfill
    \begin{subfigure}{0.41\textwidth}
        \centering
        \includegraphics[width=\linewidth]{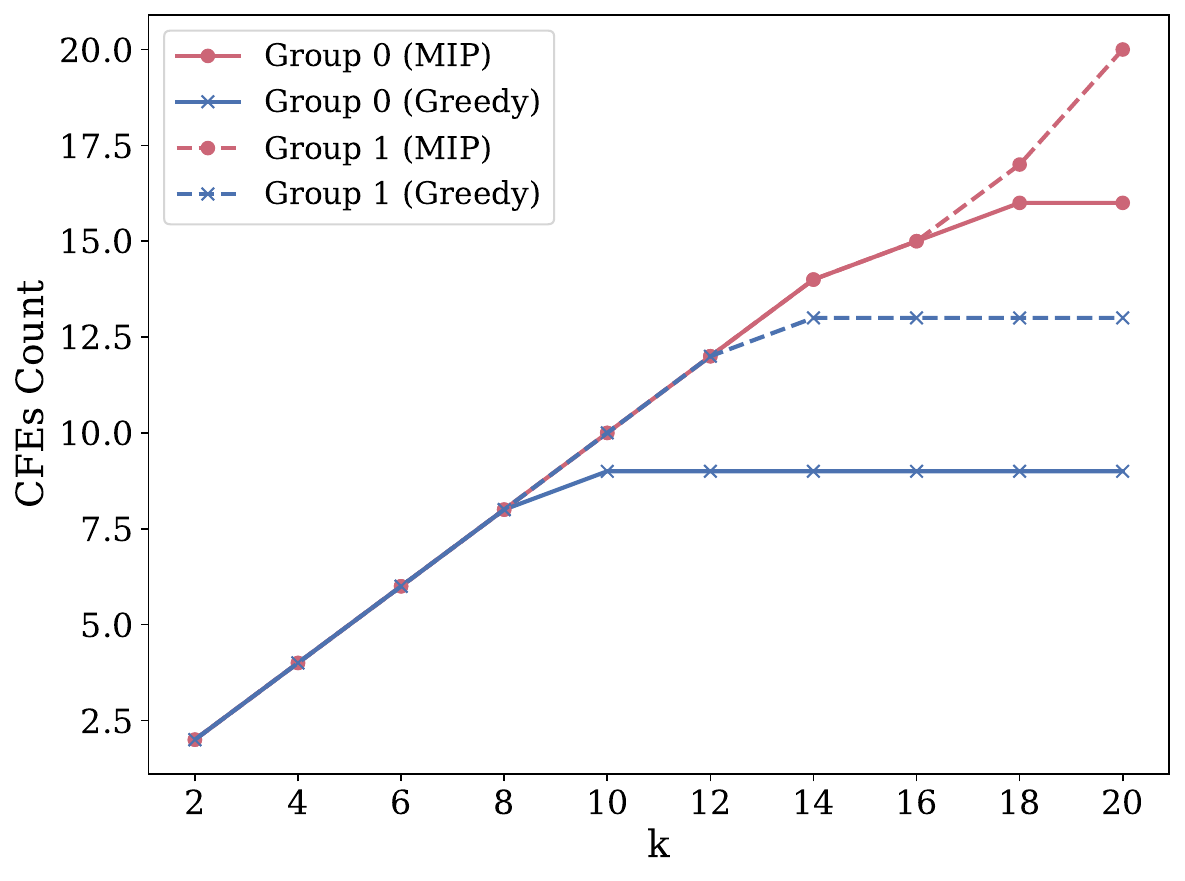} 
        \caption{Selected Counterfactuals}
    \end{subfigure}
    \caption{Comparison of Cost-Constrained Greedy and MIP algorithms.}
    \label{fig: cost_const_greedy_mip_adult}
\end{figure}

For the cost-constrained approach, Figure~\ref{fig: cost_const_greedy_mip_adult} presents the results across key metrics. Both methods achieve similar total and group-level coverage, confirming their effectiveness in selecting counterfactuals that explain the data. However, the greedy algorithm is much more computationally efficient (Figure~\ref{fig: cost_const_greedy_mip_adult}(c)) and selects a more compact set of counterfactuals (Figure~\ref{fig: cost_const_greedy_mip_adult}(d)), while the MIP continues to select unnecessary counterfactuals. Given these trade-offs, we adopt the greedy approach for its efficiency while maintaining high coverage.

\begin{figure}[h!]
    \centering   
    \begin{subfigure}{0.41\textwidth}
        \centering
        \includegraphics[width=\linewidth]{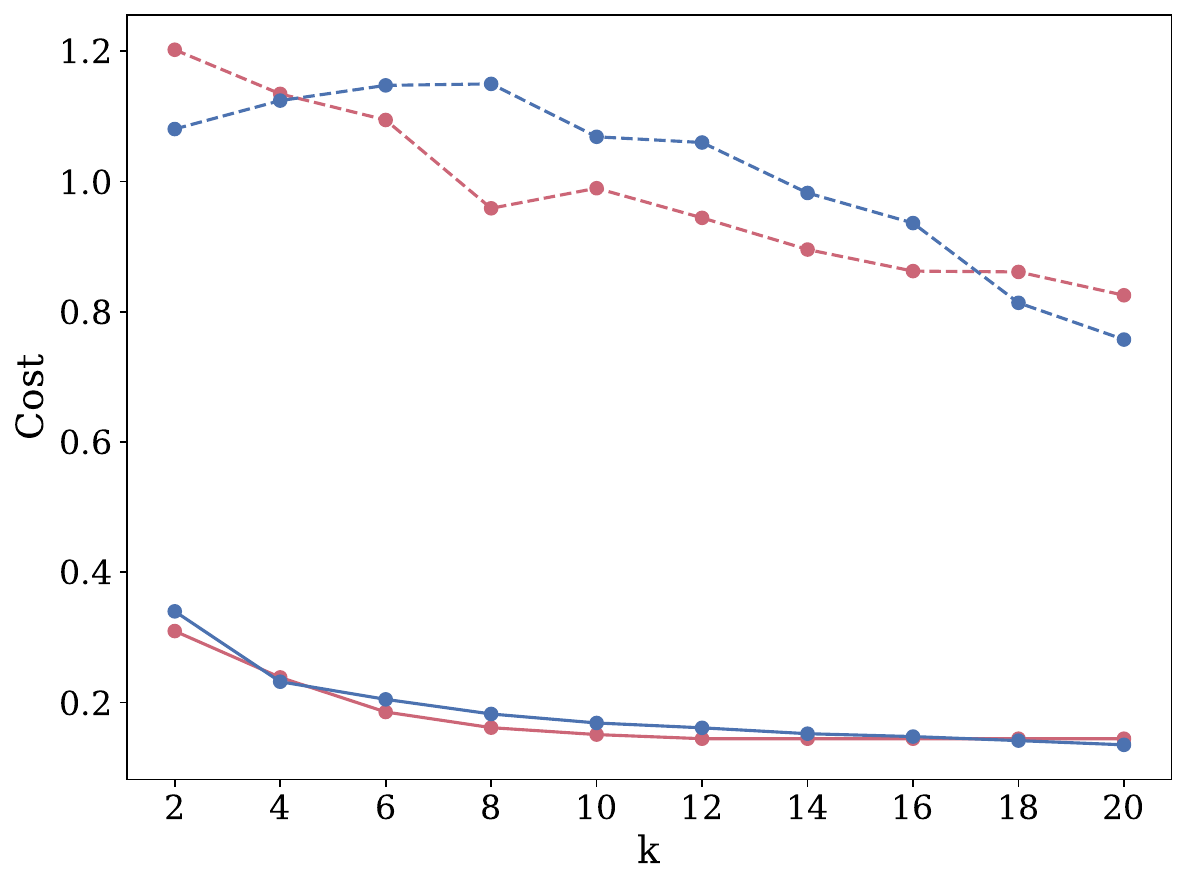}
        \caption{$c = 0.25$}
    \end{subfigure}
    \hfill
    \begin{subfigure}{0.41\textwidth}
        \centering
        \includegraphics[width=\linewidth]{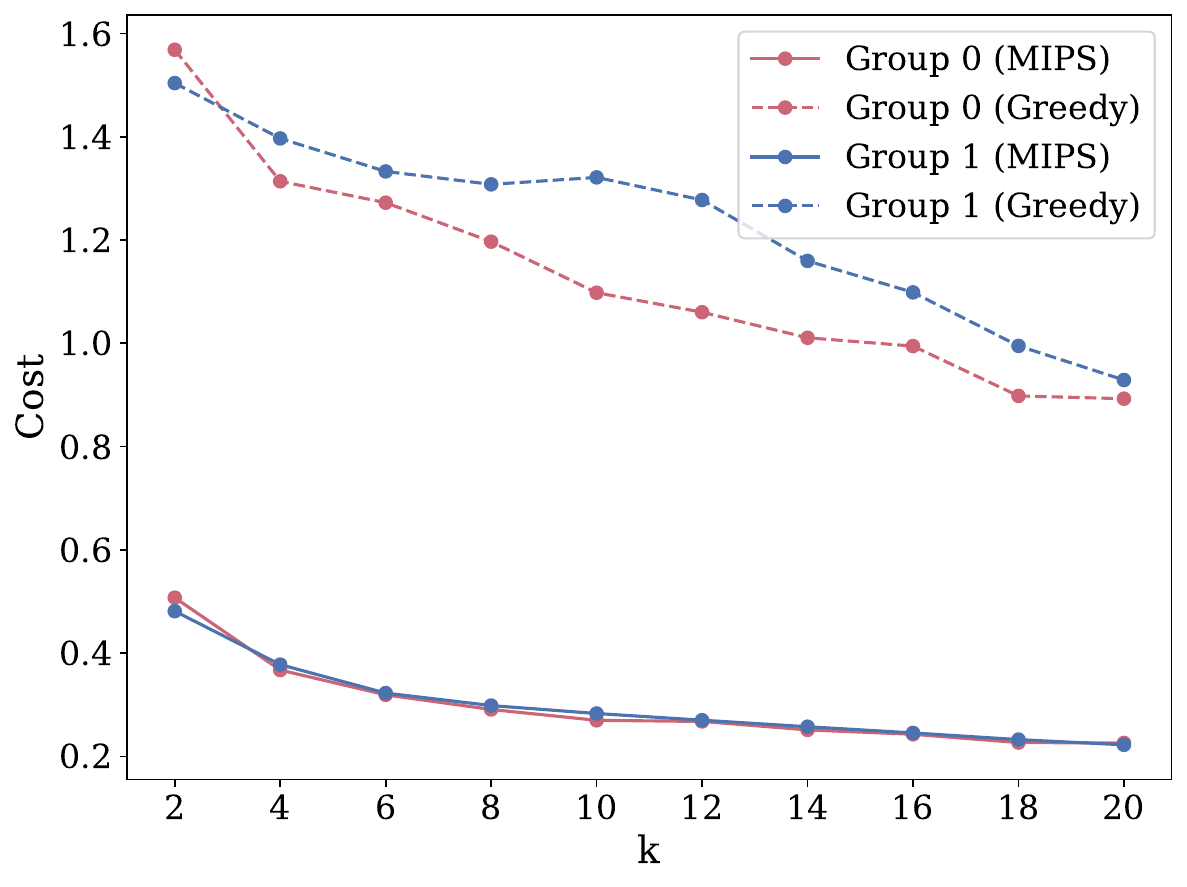}
        \caption{$c = 0.50$}
    \end{subfigure}
    \hfill
    \begin{subfigure}{0.41\textwidth}
        \centering
        \includegraphics[width=\linewidth]{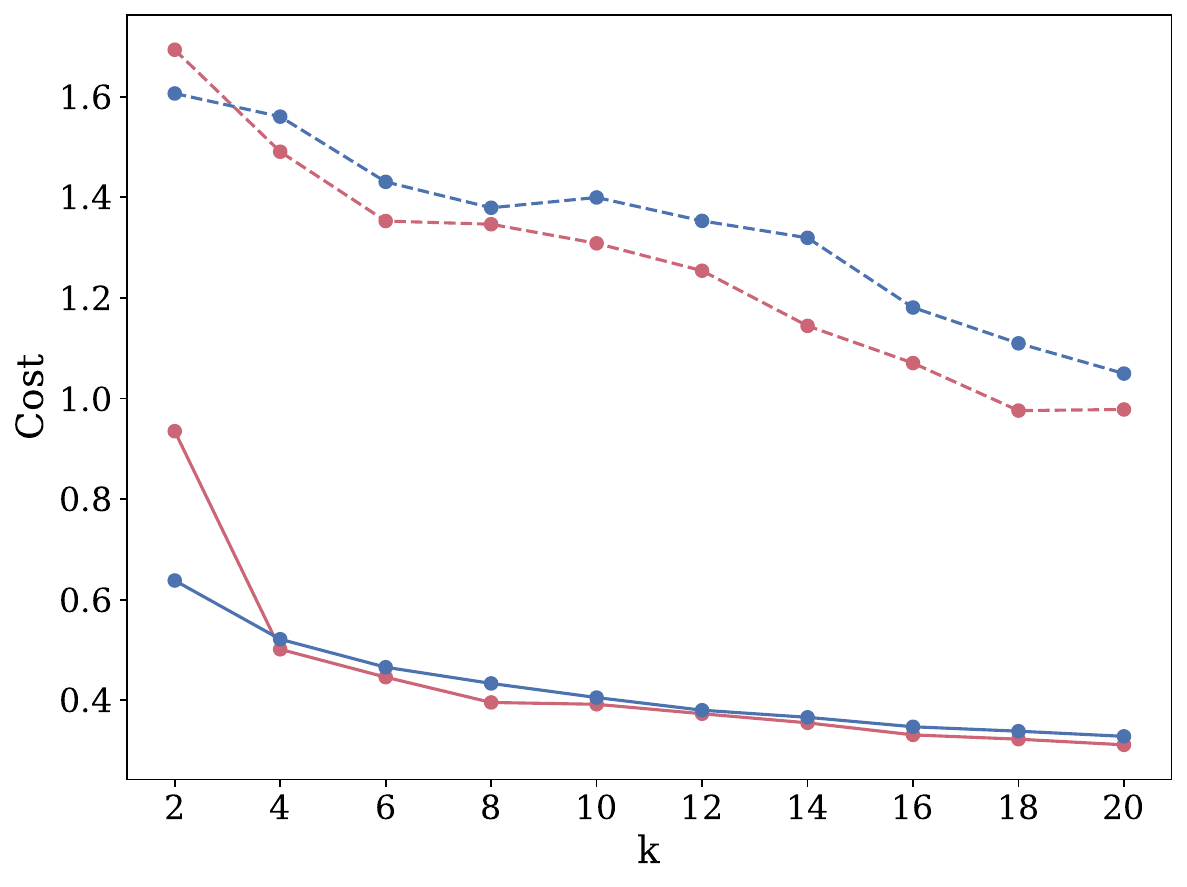} 
        \caption{$c = 0.75$}
    \end{subfigure}
    \hfill
    \begin{subfigure}{0.41\textwidth}
        \centering
        \includegraphics[width=\linewidth]{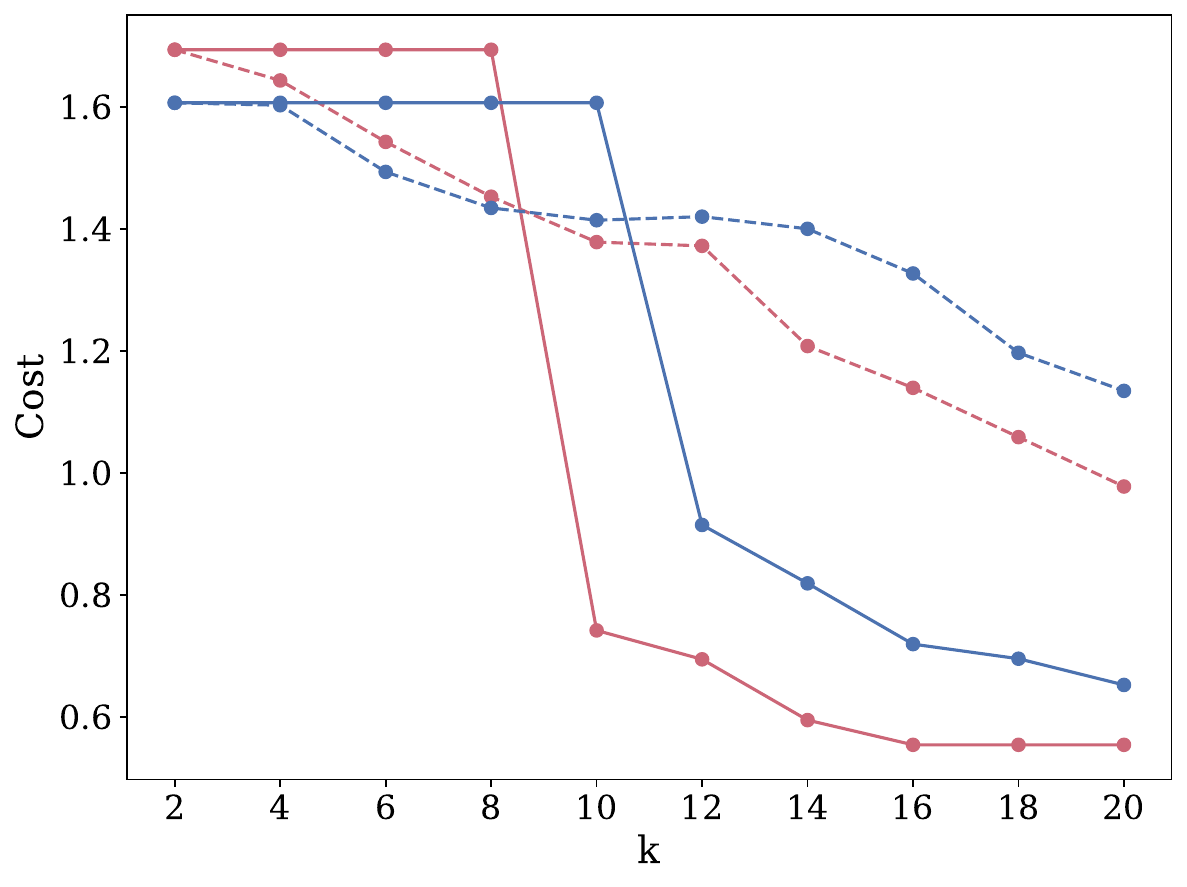} 
        \caption{$c = 1$}
    \end{subfigure}
    \caption{Comparison of Coverage-Constrained Greedy and MIP algorithms.}
    \label{fig: greedy_vs_mip}
\end{figure}

For the coverage-constrained approach, we compare how each method minimizes the maximum cost while ensuring coverage levels of $25\%, 50\%, 75\%$, and $100\%$ for $k$ from $1$ to $20$.
To account for the inherent randomness in the greedy approach, we run it 100 times and report the average results, while the MIP solver is executed with a time limit of $1800$ seconds. 
\Cref{fig: greedy_vs_mip} demonstrates that while the greedy approach is computationally efficient, MIP consistently finds lower-cost solutions, optimizing counterfactual selection under stricter cost constraints. 
This highlights the trade-off: the greedy method offers scalability at the expense of slightly higher costs, whereas MIP achieves superior solutions with significantly higher computation time.


\end{document}